\documentclass[10pt,journal,compsoc]{IEEEtran}

\ifCLASSOPTIONcompsoc
  \usepackage[nocompress]{cite}
\else
  \usepackage{cite}
\fi

\ifCLASSINFOpdf
\else
\fi

\usepackage{times}
\usepackage{epsfig}
\usepackage{graphicx}
\usepackage{amsmath}
\usepackage{amssymb}
\usepackage{footnote}
\usepackage{dsfont}
\usepackage{enumitem}

\usepackage{comment}
\usepackage{blindtext}
\usepackage{morewrites}
\usepackage{makecell}
\usepackage{pifont}%
\newcommand{\xmark}{\ding{55}}%
\newcommand{\cmark}{\ding{51}}%
\usepackage{multirow}
\usepackage{xcolor}
\usepackage{float}
\usepackage{dblfloatfix}

\newtheorem{remark}{Remark}

\begin{document}

\title{CROVIA: Seeing Drone Scenes from Car Perspective via  
Cross-View Adaptation}

\author{
Thanh-Dat~Truong$^1$~\IEEEmembership{Student~Member,~IEEE},
Chi~Nhan~Duong$^2$~\IEEEmembership{Member,~IEEE},
Ashley~Dowling$^3$, 
Son~Lam~Phung$^4$~\IEEEmembership{Senior~Member,~IEEE},
Jackson~Cothren$^5$, 
Khoa~Luu$^1$~\IEEEmembership{Member,~IEEE}
\thanks{$^{1}$Department of Computer Science and Computer Engineering, University of Arkansas;
$^{2}$Department of Computer Science and Software Engineering, Concordia University, Concordia University;
$^{3}$Department of Entomology and Plant Pathology, University of Arkansas; 
$^{4}$Faculty of Engineering and Information Sciences,
University of Wollongong; \emph{and} 
$^{5}$Department of Geosciences, University of Arkansas
Emails: \texttt{tt032@uark.edu, dcnhan@ieee.org, adowling@uark.edu, phung@uow.edu.au, jcothre@uark.edu, khoaluu@uark.edu}}
}

\IEEEtitleabstractindextext{%
\begin{abstract}
Understanding semantic scene segmentation of urban scenes captured from the Unmanned Aerial Vehicles (UAV) perspective plays a vital role in building a perception model for UAV. With the limitations of large-scale densely labeled data, semantic scene segmentation for UAV views requires a broad understanding of an object from both its top and side views. Adapting from well-annotated autonomous driving data to unlabeled UAV data is challenging due to the cross-view differences between the two data types. Our work proposes a novel Cross-View Adaptation (CROVIA) approach to effectively adapt the knowledge learned from on-road vehicle views to UAV views. First, a novel geometry-based constraint to cross-view adaptation is introduced based on the geometry correlation between views. Second, cross-view correlations from image space are effectively transferred to segmentation space without any requirement of paired on-road and UAV view data via a new Geometry-Constraint Cross-View (GeiCo) loss. Third, the multi-modal bijective networks are introduced to enforce the global structural modeling across views. Experimental results on new cross-view adaptation benchmarks introduced in this work, i.e., SYNTHIA $\to$ UAVID and GTA5 $\to$ UAVID, show the State-of-the-Art (SOTA) performance of our approach over prior adaptation methods.
\end{abstract}

}

\maketitle

\IEEEdisplaynontitleabstractindextext

\IEEEpeerreviewmaketitle

\IEEEraisesectionheading{\section{Introduction}\label{sec:intro}}

Unmanned Aerial Vehicles (UAV), colloquially called drones, have been widely adopted in various practical applications such as autonomous aerial flying, 3D map generation, multi-object or wildlife tracking, disaster management, precision agriculture, etc. UAV images and videos provide complement high altitude aerial and satellite images with higher resolution details for these applications due to their low altitude and flexible flight paths. 
Annotating high-resolution UAV images is a time-consuming and costly process. At present, there exist many large-scale autonomous driving datasets captured from on-road vehicles, e.g., Cityscapes \cite{cordts2016cityscapes}, SYNTHIA \cite{Ros_2016_CVPR}, GTA5 \cite{Richter_2016_ECCV}. They have been studied for many years and are well-annotated, especially for densely class prediction tasks such as semantic scene segmentation. As these datasets share objects of interest to UAV data, leveraging the knowledge from on-road vehicles to UAV ones can significantly benefit the learning process to reuse large-scale annotations and save efforts of annotating UAV images manually. However, on-road images captured on the ground differ from those captured from UAV views, where only the top of objects is visible. It leaves a challenging domain gap between the two views. Thus, although there have been many works on Unsupervised Domain Adaptation (UDA) in semantic segmentation, there are limited studies in cross-view adaptation.

\begin{figure}[!t]
    \centering
    \includegraphics[width=0.50\textwidth]{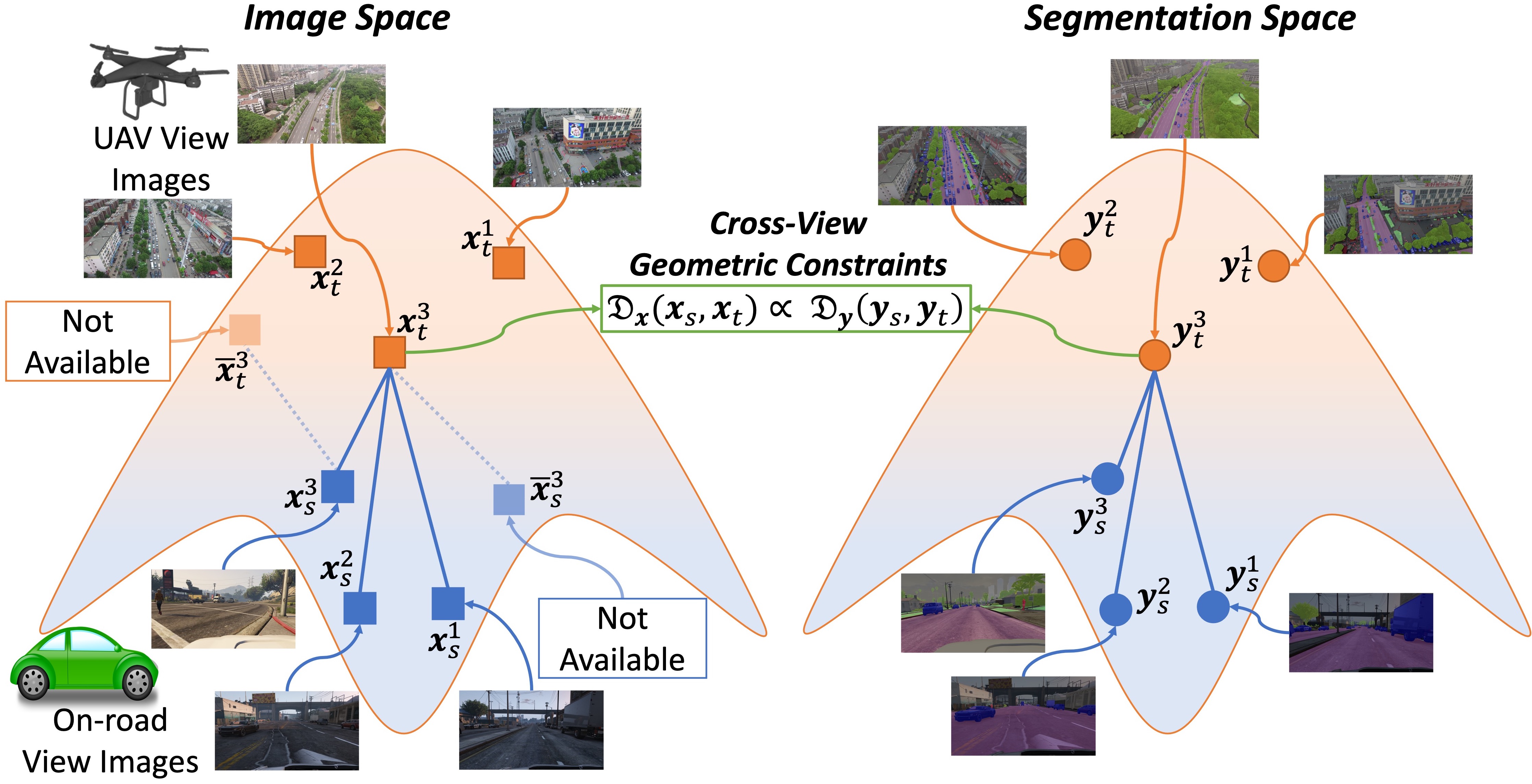}
    \caption{
    \textbf{The Geometric Constraint Across Views between Images and Segmentation Maps for Cross-View Adaptation}. The paired data between two views are not available, e.g., $\mathbf{\bar{x}}_t$ (or $\mathbf{\bar{x}}_s$) is a corresponding image of $\mathbf{x}_s$ (or $\mathbf{x}_t$) in the opposite views, which is not available. %
    Our proposed \textbf{\textit{GeiCo}} loss is proven to impose the cross-view geometric constraints via unpaired samples.
    }
    \label{fig:car_drone_view}
\end{figure}

\begin{table*}[t]
\small
\centering
\caption{Comparisons in the properties between our proposed CROVIA approach and prior domain adaptation methods. Ent: Entropy Minimization, $\ell_{adv}$: Adversarial Loss, $\ell_{depth}$: Depth Huber Loss, $\ell_{bml}$: Bijective Maximum Likelihood Loss, $\ell_{pce}$: Cross Entropy Loss with Pseudo Labels, $\ell_{focal}$: Focal Loss, $\ell_{kwd}$: Knowledge Distillation Loss, $\ell_{feat}$: Feature Distance Loss.}
\begin{tabular}{c|c|c|c|c|c}
\hline
\textbf{Methods} & \textbf{\begin{tabular}[c]{@{}c@{}}Geometric \\  Aware\end{tabular}} & \textbf{\begin{tabular}[c]{@{}c@{}}Topology Preserving \\ Aware\end{tabular}} & \textbf{Learning Approach}                                                      & \textbf{\begin{tabular}[c]{@{}c@{}}Cross-view \\ Adaptation\end{tabular}} & \textbf{\begin{tabular}[c]{@{}c@{}}Structural Learning \\ Strategy \end{tabular}}                           \\ 
\hline

AdvEnt\cite{vu2019advent}  & \xmark & \xmark & Ent + $\ell_{adv}$ & \xmark & Weak Indication by $\ell_{adv}$ \\ 
\hline

DADA\cite{vu2019dada} & \xmark & \xmark & $\ell_{adv} + \ell_{depth}$ & \xmark & Depth-Aware Structure \\ 
\hline

IntraDA\cite{pan2020unsupervised}  & \xmark & \xmark & Ent + Curriculum Training& \xmark & Weak Indication $\ell_{Adv}$ \\ 
\hline

BiMaL\cite{truong2021bimal} & \xmark & \xmark & $\ell_{bml}$ & \xmark & Distribution Modeling \\ 
\hline

SAC\cite{Araslanov:2021:DASAC} & \xmark & \xmark & $\ell_{pce} + \ell_{focal}$ & \xmark & Augmentation Consistency \\ 
\hline

ProDA\cite{zhang2021prototypical} & \xmark & \xmark & $\ell_{pce} + \ell_{kwd}$ & \xmark & Augmentation Consistency \\ 
\hline

DAFormer\cite{daformer} & \xmark & \xmark & $\ell_{pce} + \ell_{feat}$ & \xmark & \begin{tabular}[c]{@{}c@{}} Perceptual Feature by $\ell_{feat}$ \end{tabular} \\
\hline

\textbf{Our CROVIA} & \cmark & \cmark
& \textbf{\begin{tabular}[c]{@{}c@{}}Geometry Constraint \\ Cross-View Loss\end{tabular}} & \cmark & \textbf{\begin{tabular}[c]{@{}c@{}}Cross-View Multi-modal \\  Bijective Networks\end{tabular}} \\ 
\hline
\end{tabular}
\label{tab:summary1}
\end{table*}

Typically, UDA methods can be divided into adversarial learning and self-supervised learning. The former methods focus on minimizing the distribution discrepancy of the deep representations between the source and target domains via mean discrepancy \cite{ganin2015unsupervised, long2015learning}, adversarial loss \cite{chen2018road, hoffman18a, hong2018CVPR, tsai2018learning, tzeng2017adversarial}, maximum likelihood loss \cite{truong2021bimal}, contrastive learning  \cite{Yue_2021_CVPR, kang2019contrastive}. The latter methods utilize the pseudo labels and training techniques \cite{Araslanov:2021:DASAC, daformer, zhang2021prototypical}.
Both approaches have shown their potential to adapt to different environmental variations, e.g., lighting or geographical domain shifts.

\begin{figure*}[!t]
    \centering
    \includegraphics[width=1.0\textwidth]{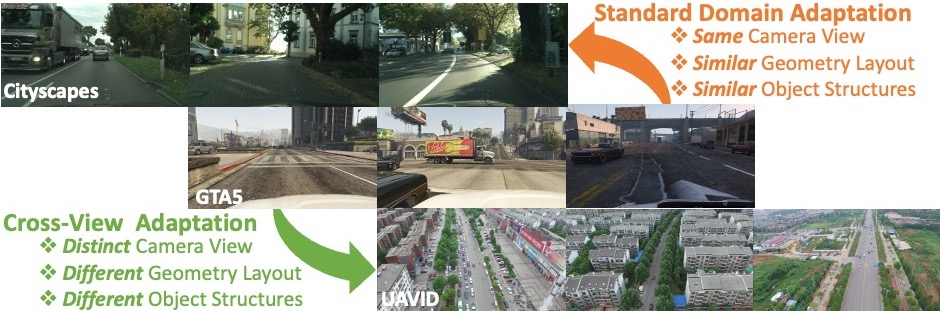}
    \caption{The Comparison between Domain Adaptation (GTA5 $\to$ Cityscapes) and Cross-View Adaptation (GTA5 $\to$ UAVID).}
    \label{fig:compare_da_ca}
\end{figure*}

However, there are several challenges to applying these methods directly to UAV data. First, \textbf{\textit{cross-view differences}} between on-road and UAV data are usually significant, and the position knowledge of each view is not explicitly encoded in Convolutional Neural Networks (CNNs) or Transformers. 
Moreover, the distributions of the two domains are quite different. 
For example, cars appear in their side view from the street, but they are visible only in their top view from the UAV.
Therefore, minimizing their deep representations is not a straightforward optimization problem. Second, the \textit{geometric layout, i.e., topology structure,}  of these two views differs significantly.
For example, as shown in Fig. \ref{fig:compare_da_ca}, objects' shapes and structures, e.g., cars, trucks, or trees, are similar between source and target domains in prior methods. However, they are quite different between on-road and UAV images. Moreover, the relative structures or topological constraints among objects in on-road datasets are different for UAV images, e.g., trees and buildings are on two sides of the road.
Thus, a model learned on an on-road vehicle dataset cannot generalize well on UAV images.

\noindent
\textbf{Contributions of this Work:}  This work proposes a novel geometry-constraint Cross-View Adaptation (CROVIA) approach to effectively adapt the knowledge learned from on-road vehicle views to UAV views. Our contributions are four-fold. 
First, by analyzing the limitations of prior UDA methods in the cross-view settings, a new Geometry-constraint Cross-view (GeiCo) metric is presented for cross-view adaptation on unpaired data (as shown in Fig. \ref{fig:car_drone_view}).
Second, this metric is further derived into the new GeiCo loss function to incorporate into the deep neural network to improve the adaptation process from on-road to UAV views. Third, in contrast to prior work, a new multi-modal bijective network-based approach is presented to enforce the global structure embedding process across views in GeiCo loss. Finally, a new benchmark for cross-view adaptation is presented. Experimental results have shown the proposed CROVIA approach outperforms all prior domain adaptation methods. Table \ref{tab:summary1} summarizes the difference between the proposed CROVIA and prior methods. To the best of our knowledge, this work is \textit{one of the first studies} addressing the cross-view adaptation in semantic scene segmentation.

\section{Related Work}

\noindent
\textbf{Semantic Scene Segmentation.}
Convolutional Neural Networks \cite{long2015fully, chen2018deeplab, lin2017refinenet} have shown their performance in semantic segmentation applications in both general \cite{lin2017refinenet, chen2018deeplab} and UAV images \cite{UNetFormer}. The performance of CNNs is further improved by using multi-level features \cite{long2015fully, chen2018deeplab}, dilated convolution \cite{chen2018deeplab, DBLP:journals/corr/YuK15}, or spatial pyramid pooling  \cite{lin2017refinenet, pohlen2017full}. Recent studies promoted the performance of segmentation models by utilizing Transformer networks \cite{xie2021segformer, UNetFormer}. However, training supervised models require a large amount of annotated data.

\noindent
\textbf{Unsupervised Domain Adaptation} plays a role in alleviating the demand for large amounts of annotated data. Adversarial learning \cite{tzeng2017adversarial, chen2018road, chen2017no, tsai2018learning, vu2019advent, vu2019dada} and self-supervised learning \cite{daformer, Araslanov:2021:DASAC, zhang2021prototypical} are  two primary approaches to UDA.

\noindent
\textbf{Adversarial Learning Methods} are the preferred UDA approaches for semantic scene segmentation where the model is optimized simultaneously on the source and target domains
within an adversarial framework. 
Hoffman et. al. \cite{hoffman2016fcns} introduced the first adversarial learning-based approach to UDA in semantic segmentation. Chen et. al. \cite{chen2018road} presented target-guided distillation loss with a spatial-aware model in the UDA framework. Assuming sharing structures between source and target domains, \mbox{Tasi et. al.} \cite{tsai2018learning, tsai2019domain} presented an adversarial training to model the distributions across domains.
Several authors \cite{zhu2017unpaired, murez2018CVPR, hoffman18a} utilized an image translation approach to translate a source to a novel target. Zhu et. al. \cite{zhu2018ECCV} proposed a conservative loss to penalize easy and hard source samples. SPIGAN \cite{lee2018spigan} and DADA \cite{vu2019dada} utilized privileged depth information to learn a depth-aware model. Recent studies have used entropy minimization-based approaches in UDA. Vu et. al. \cite{vu2019advent} first presented an adversarial entropy minimization approach in semantic scene segmentation under the UDA setting. 
\cite{pan2020unsupervised, 10.1145/3474085.3475174} presented an entropy-based curriculum adaptation framework including two phases, i.e., inter- and intra-domain adaptation. Truong et. al. \cite{truong2021bimal} generalized entropy minimization by introducing the bijective maximum likelihood loss.

\noindent
\textbf{Self-supervised Approaches } 
have gained SOTA performance in semantic scene segmentation \cite{Araslanov:2021:DASAC, daformer, zhang2021prototypical} in recent years. 
In self-supervised methods, a new model is trained on the target domain using pseudo labels produced from predictions of a trained model on the source domain. 
Zou et. al. \cite{zou2018unsupervised} presented a class-balanced self-training method for UDA in semantic scene segmentation. 
Araslanov et. al. \cite{Araslanov:2021:DASAC} developed a self-supervised augmentation consistency framework to evolve the pseudo labels without additional training rounds.
Zhang et. al. \cite{zhang2021prototypical} 
introduced a method to online correct the soft pseudo labels and utilize knowledge distillation to boost the performance of models.
Hoyer et. al. \cite{daformer} improved the performance of the domain adaptation via a new Transformer-based backbone and training recipe and then further improved by a context-aware high-resolution domain-adaptive framework \cite{hoyer2022hrda}.

\noindent
\textbf{Cross-View Learning}
Several works have exploited cross-view learning for the geo-localization applications \cite{zhu2022transgeo, Toker_2021_CVPR, shi2020where, NIPS2019_9199}. They aim to learn the joint embedding spaces of street-view and aerial-view images.
They applied a predefined polar transform \cite{ shi2020where, NIPS2019_9199} on the aerial-view images so that the transformed aerial images have a similar geometric layout to the street-view images.
Other studies \cite{9010757, Toker_2021_CVPR} utilized the conditional generative adversarial networks to synthesize street-view images from corresponding aerial-view images.
Meanwhile, \cite{zhu2022transgeo} learns the correlation between street-view and aerial-view images using the self-attention mechanism.
\cite{NoVA} a cross-view adaptation approach between car and truck views. 
Despite the change of camera positions between cars and trucks, The change of views is not significant as our problem where the views change from cars to drones. Also, \cite{NoVA} requires depth labels and a transformation function between views during training. 
\cite{ADeLA} presented a cross-viewpoint adaptation but has a different focus from ours. It requires 3D models of scenes to create pairs of images between views.
\cite{SceneAdapt} introduced an adversarial approach trained on the proposed synthetic dataset. 
Although the dataset in \cite{SceneAdapt} contains multi-views, these images were taken from the same altitude of a 
camera position with different pitch and yaw angles. 
Meanwhile,  in our problem, the camera views are placed at different altitudes, i.e., a forward view on the road and a slanted view at a high altitude, which results in a significant difference in the scene structures. 
To the best of our knowledge, cross-view adaptation learning in semantic scene segmentation has yet to be widely exploited. Therefore, we present a novel approach to cross-view adaptation in semantic segmentation.

\noindent
\textbf{Bijective Networks}
NICE \cite{dinh2015nice} and RealNVP \cite{dinh2017density} presented the initial work of normalizing the flow-based model by introducing the invertible bijective transformation. The transformation is designed as an affine coupling layer which is a tractable form of computing the Jacobian determinants. Then, RealNVP \cite{dinh2017density} introduced the multi-scale architecture to model for the large image size. Batch normalization and weight normalization have also been introduced to improve training efficiency.
Germain et. al. \cite{made} presented an autoregressive autoencoder that can estimate the distributions and be able to compute the Jacobian determinant easily. \cite{mask_autoregessive} introduces the masked autoregressive flow that the network is designed in an autoregressive manner.
Kingma et. al.  \cite{glow} introduce the intertible $1 \times 1$ convolutions to learn the permutation matrix instead of fixing a permutation as in RealNVP \cite{real_nvp}. 
 Hoogeboom et. al. \cite{dxd_invertible_conv} proposed an invertible $n \times n$ convolution generalized from the $1 \times 1$ convolution.

\section{Our Cross-View Adaptation Approach} \label{sec:cross_view_adapt}

This section first reviews standard UDA and its limitations in cross-view settings. Then, the new cross-view adaptation is presented via the cross-view geometric constraints.

\subsection{Unsupervised Cross-View Adaptation}

Let $\{\mathbf{x}_s, \mathbf{\hat{y}}_s\}$ be a pair of an image $\mathbf{x}_s \in \mathcal{X}_s$ and its segmentation label $\mathbf{\hat{y}}_s \in \mathcal{Y}$ captured from an on-road view. $\mathcal{X}_s \subset \mathbb{R}^{H \times W \times 3}$ denotes the image space; $H$ and $W$ are the height and width of $\mathbf{x}_s$ respectively. $\mathcal{Y} \subset \mathbb{R}^{H \times W \times C}$ is the segmentation space where $C$ is the number of classes. 
Similarly, $\mathbf{x}_t \in \mathcal{X}_t$ is an image captured from a UAV view.  
We define $F: \mathcal{X}_s \cup \mathcal{X}_t \to \mathcal {Y}$ as a deep segmentation network with parameters $\theta$ that maps an input image to its corresponding segmentation map $\mathbf{y}$, i.e., $\mathbf{y}_s = F_\theta(\mathbf{x}_s)$ and $\mathbf{y}_t = F_\theta(\mathbf{x}_t)$.
Given $\mathbf{x}_s$ and $\mathbf{\hat{y}}_s$ drawn from an on-road data distribution $p_s(;)$, i.e., $\mathbf{x}_s, \mathbf{\hat{y}}_s \sim p_s(\mathbf{x}_s, \mathbf{\hat{y}}_s)$, 
and $\mathbf{x}_t$ drawn from an UAV data distribution $p_t(;)$, i.e., $\mathbf{x}_t \sim p_t(\mathbf{x}_t)$, the cross-view adaptation can be formed as in Eqn. \eqref{eqn:general_obj}.
\begin{equation} \label{eqn:general_obj}
\scriptsize
    \arg\min_{\theta} \Big[\mathbb{E}_{\mathbf{x}_s, \mathbf{\hat{y}}_s \sim p_s(\mathbf{x}_s, \mathbf{\hat{y}}_s)} \big[\mathcal{L}_{s}(\mathbf{y}_s, \hat{\mathbf{y}}_s)] 
    + \mathbb{E}_{\mathbf{y}_t \sim p_t(\mathbf{y}_t)} [\mathcal{L}_{t}(\mathbf{y}_t \big| p_s)\big]\Big]\\
\end{equation}
where $\mathcal{L}_s$ and $\mathcal{L}_t$ are the losses defined on \textit{the on-road} and \textit{the UAV domains}, respectively. As the groundtruth of the on-road domain is available, $\mathcal{L}_s$ can be defined as the cross-entropy loss. $\mathcal{L}_t$ in our work will be detailed in Sec. \ref{sec:learn_on_unpair}.

In prior work, \cite{vu2019advent, Araslanov:2021:DASAC, daformer, SDHV18}, the adaptation setting is employed in the context of environmental changes, e.g., simulator to real \cite{vu2019advent, Araslanov:2021:DASAC, daformer} or weather changes \cite{SDHV18}. The camera views and spatial layouts are usually similar in this setting and result in a shared geometric layout in segmentation maps of the two domains.
Thus, the distribution $p_t$ can be approximated by $p_s$, and  $\mathcal{L}_t$ can be defined as an adversarial loss  \cite{truong2021bimal, tsai2019domain, vu2019advent} or self-supervised loss \cite{Araslanov:2021:DASAC, daformer}.
However, in the cross-view adaptation setting, camera view changes (from on-road vehicle to UAV view) bring larger differences in geometric layout and topological structures of both RGB appearances and segmentation maps; and make the distribution approximation as prior work inefficient. 
Hence, direct adoption of prior approaches to cross-view adaptation cannot bring potential improvements (see Sec. \ref{sec:quantitative_result} for comparison with UDA approaches) to predictions of UAV data.

To effectively address cross-view adaptation, two properties should be considered:
(1) \textit{\textbf{geometric correlations between views}} so that the view changes from RGB images can be effectively transferred to segmentation for adaptation on UAV views, and  
(2) \textit{\textbf{global structure learning}} so that global topology structures of UAV views are enforced during the adaptation process.

\subsection{Geometric Correlation Across Views}

As segmentation labels of UAV data are not accessible during training, their distributions cannot be directly modeled. This section first proposes to model the geometric correlations across views in image space. Then, this knowledge is transferred to the segmentation space to maintain the geometric consistency for $F_\theta$.
Given an on-road view image $\mathbf{x}_s$, $\mathbf{\bar{x}}_t$ is defined as its corresponding image captured from UAV view with a segmentation map $\mathbf{\bar{y}}_t$. Let $\mathcal{D}_{\mathbf{x}} (\mathbf{x}_s, \mathbf{\bar{x}_t})$ and $\mathcal{D}_{\mathbf{y}}(\mathbf{y}_s, \mathbf{\bar{y}_t})$ be the metrics measure the correlations between images (i.e. $\mathbf{x}_s$ and $\mathbf{\bar{x}}_t$) and segmentations (i.e. $\mathbf{y}_s$ and $\mathbf{\bar{y}}_t$). 
\begin{remark} \label{remark:linear_trans}
\textbf{The Geometric Transformation Between Camera Views.}
Since $\mathbf{x}_s$ and $\mathbf{\bar{x}}_t$ are captured from two camera positions of the same scene, the geometric transformation from $\mathbf{x}_s$ to $\mathbf{\bar{x}}_t$ can be modeled via a transformation matrix $\mathbf{T}_{s \to t}$ as in Eqn. \eqref{eqn:warp_func}.
\begin{equation} \label{eqn:warp_func}
     \mathbf{\bar{x}}_t = \mathcal{W}(\mathbf{x}_s, \mathbf{T}_{s \to t})
\end{equation}
where $\mathcal{W}:\mathcal{X}_s \times \mathbb{R}^{3 \times 3} \to \mathcal{X}_t$ denotes a warping function, i.e. $\mathbf{\bar{x}}_t (\mathbf{\bar{p}}) = \mathbf{x}_s (\mathbf{p})$, $\mathbf{\bar{p}} = \mathbf{T}_{s \to t} \times \mathbf{p}$. $\mathbf{p}$ and $\mathbf{\bar{p}}$ are the pixel locations in $\mathbf{x}_s$ and $\mathbf{\bar{x}}_t$, respectively.
\end{remark}

\begin{remark} \label{remark:equi_trans}
\textbf{The Equivalent Transformation Between Image and Segmentation.}
As RGB images and segmentation maps are pixel-wised corresponding, the same transformation can be adopted for segmentation maps as in Eqn. \eqref{eqn:linear_xs_to_xt}.
\begin{equation} \label{eqn:linear_xs_to_xt}
     \mathbf{\bar{y}}_t = \mathcal{W}(\mathbf{y}_s, \mathbf{T}_{s \to t})
\end{equation}
\end{remark}
In practice, the warping function $\mathcal{W}$ can be derived from $\mathbf{T}_{s \to t}$ and presented in the form of a permutation matrix $\mathbf{W}_{s \to t} \in \mathbb{R}^{HW \times HW}$. $\mathbf{\bar{x}}_t$ and $\mathbf{\bar{y}}_t$ are then reformulated as,
\begin{equation} \label{eqn:xt_yt_reformulation}
\begin{split}
    \mathbf{\bar{x}}_t &= \mathbf{W}_{s \to t} \times \mathbf{x}_s 
    \\
     \mathbf{\bar{y}}_t &= \mathbf{W}_{s \to t} \times \mathbf{y}_s
\end{split}
\end{equation}
In Eqn. \eqref{eqn:xt_yt_reformulation}, the geometry changes from $\mathbf{x}_s$ to $\mathbf{\bar{x}}_t$, 
and $\mathbf{y}_s$ to $\mathbf{\bar{y}}_t$ 
are mainly reflected in $\mathbf{W}_{s \to t}$. 
Thus, the correlations between $\mathcal{D}_{\mathbf{x}} (\mathbf{x}_s, \mathbf{\bar{x}_t})$ and $\mathcal{D}_{\mathbf{y}}(\mathbf{y}_s, \mathbf{\bar{y}_t})$ rely on both $\mathbf{W}_{s \to t}$ and the difference between $\mathbf{x}_s$ and $\mathbf{y}_s$. The proportion in the relationships between $\mathcal{D}_{\mathbf{x}} (\mathbf{x}_s, \mathbf{\bar{x}_t})$ and $\mathcal{D}_{\mathbf{y}}(\mathbf{y}_s, \mathbf{\bar{y}_t})$ can be defined as in Eqn. \eqref{eqn:constraint_dx_dy}.
\begin{equation} \label{eqn:constraint_dx_dy}
\begin{split}
   \mathcal{D}_{\mathbf{x}}(\mathbf{x}_s, \mathbf{\bar{x}}_t)  &\propto  \mathcal{D}_{\mathbf{y}}(\mathbf{y}_s, \mathbf{\bar{y}}_t) 
   \\
    \Leftrightarrow \quad \mathcal{D}_{\mathbf{x}}(\mathbf{x}_s, \mathbf{\bar{x}}_t)  &= \alpha \mathcal{D}_{\mathbf{y}}(\mathbf{y}_s, \mathbf{\bar{y}}_t)
\end{split}
\end{equation}
\subsection{Geometry Constraint Cross-view (GeiCo) Metric on Unpaired Data}
\label{sec:learn_on_unpair}

\begin{figure*}
    \centering
    \includegraphics[width=0.9\textwidth]{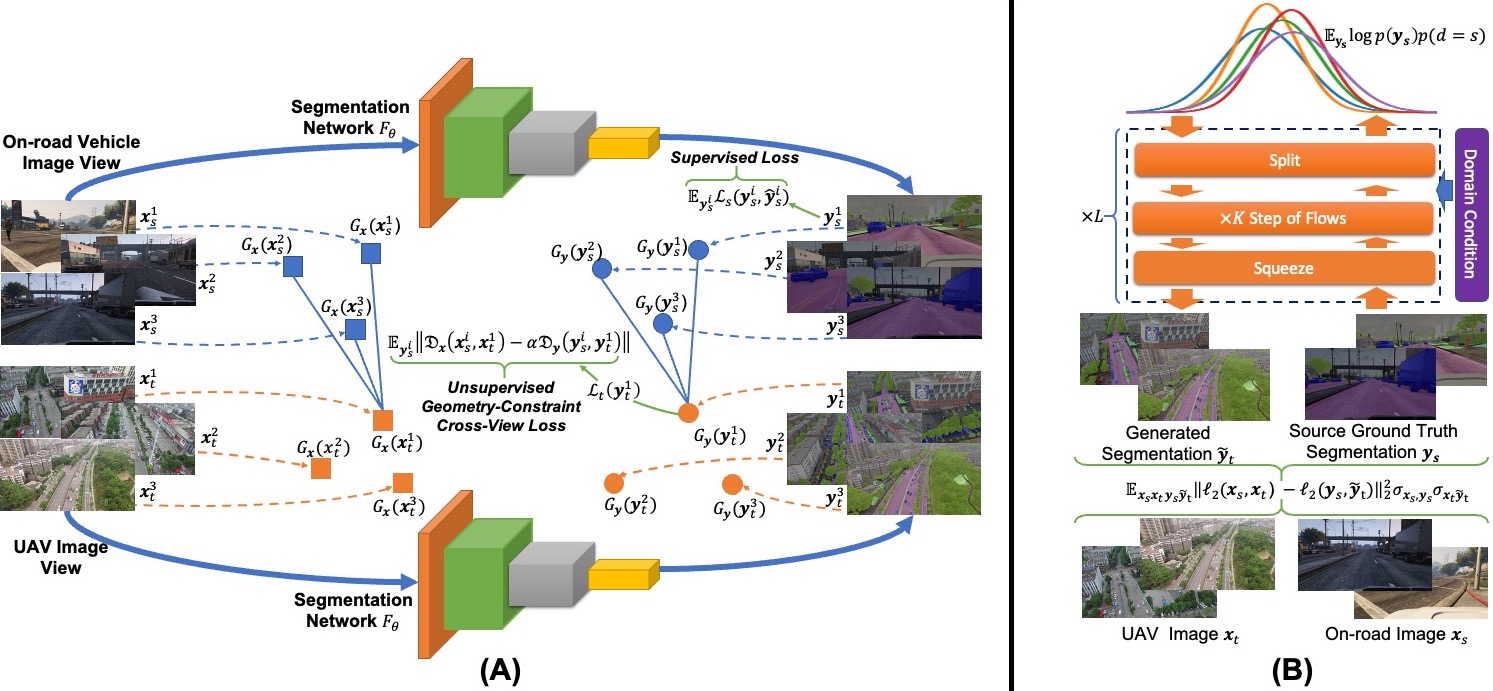}
    \caption{\textbf{(A) The Proposed CROVIA Framework.} The input images $(\mathbf{x}_s, \mathbf{x}_t)$ in both views are first forwarded to the segmentation $F_\theta$. Then, the predictions of the on-road vehicle view are imposed by the supervised loss with ground truths, i.e., $\mathcal{L}_s(\mathbf{y}_s, \mathbf{\hat{y}}_s)$. Meanwhile, the predictions of the UAV view are penalized by the unsupervised Geometry-Constraint Cross-View loss, i.e., $\mathcal{L}_t(\mathbf{y}_t)$.
    \textbf{(B) The Proposed Multi-Model Bijective Networks with Domain Condition.} $G_\mathbf{y}$ is learned by maximizing the likelihood of on-road-view segmentation maps in parallel with optimizing the regularizer $\mathcal{R}(\mathbf{\widetilde{y}}_t)$ of generated segmentation maps $\mathbf{\widetilde{y}}_t$.
    }
    \label{fig:proposed_framework}
\end{figure*}
Eqn. \eqref{eqn:constraint_dx_dy} defines a necessary condition to explicitly model the geometry constraints between camera views. Then, Eqn. \eqref{eqn:general_obj} can be formed as,
\begin{equation} 
\label{eqn:obj_with_cond_pair}
\small
\footnotesize
\begin{split}
     &\arg\min_{\theta} \Big[\mathbb{E}_{\mathbf{x}_s, \mathbf{\hat{y}}_s \sim p_s(\mathbf{x}_s, \mathbf{\hat{y}}_s)} \mathcal{L}_{s}(\mathbf{y}_s, \hat{\mathbf{y}}_s)
    + \mathbb{E}_{\mathbf{y}_t \sim p_t(\mathbf{y}_t)} \mathcal{L}_{t}(\mathbf{y}_t \big| p_s )\Big]\\
    &s.t. \quad\quad\quad\quad\quad  \mathcal{D}_{\mathbf{x}}(\mathbf{x}_s, \mathbf{\bar{x}}_t)  = \alpha \mathcal{D}_{\mathbf{y}}(\mathbf{y}_s, \mathbf{\bar{y}}_t)
\end{split}
\end{equation}
Solving Eqn. \eqref{eqn:obj_with_cond_pair} in ideal cases is straightforward if the pair of  $\mathbf{x}_s$ and $\mathbf{\bar{x}}_t$ are available. 
The unsupervised loss $\mathcal{L}_t$ can be defined as a regularizer of this constraint, i.e., $\mathcal{L}_t(\mathbf{y}_t | p_s) = ||\mathcal{D}_{\mathbf{x}}(\mathbf{x}_s, \mathbf{\bar{x}}_t)  - \alpha \mathcal{D}_{\mathbf{y}}(\mathbf{y}_s, \mathbf{\bar{y}}_t)||_2^2$. 
However, in practice, the pair data between on-road and UAV views are inaccessible as images of these two views are often collected independently. Hence, solving Eqn. \eqref{eqn:obj_with_cond_pair} with unpaired data is challenging but needed in this problem.

Instead of solving Eqn. \eqref{eqn:obj_with_cond_pair} with pair data, we consider all unpaired samples $(\mathbf{x}_s, \mathbf{x}_t)$ $\left(\text{and }(\mathbf{y}_s, \mathbf{y}_t)\right)$ between two views. 
In addition, we assume that the correlation between images (and segmentation maps) between two views is bounded by a certain threshold $\beta$, i.e.,  $\forall \mathbf{x}_s \mathbf{x}_t: \mathcal{D}_\mathbf{x}(\mathbf{x}_s, \mathbf{x}_t) \leq \beta$ (similar for $\mathcal{D}_{\mathbf{y}}$). 
This bounded constraint means that the distribution shifts across views are constrained by a threshold $\beta$ to guarantee the generalizability of the model against the distribution of data expanded from the on-road to the UAV view.
Intuitively, although the cross-view pair samples are not available, the constraints between two views can be modeled by imposing the topological constraint among unpaired samples. 
In particular, our \textbf{\textit{geometry constraint cross-view metric}} $\mathcal{L}_t$ between unpaired samples can be defined as in Eqn. \eqref{eqn:loss_for_lt}.
\begin{equation}\label{eqn:loss_for_lt} 
\begin{split}
    \mathcal{L}_t(\mathbf{y}_t|p_s) = \mathbb{E}_{\mathbf{x}_s \sim p_s(\mathbf{x}_s)} ||\mathcal{D}_{\mathbf{x}}(\mathbf{x}_s, \mathbf{x}_t) - 
    \alpha\mathcal{D}_{\mathbf{y}}(\mathbf{y}_{s}, \mathbf{y}_{t})||_2^2
\end{split}
\end{equation}
where $||\cdot||_2$ is the $\ell_2$ norm.
Hence, optimizing Eqn. \eqref{eqn:loss_for_lt} \textbf{\textit{does NOT require the demand of pair data}}.
Importantly, it can be proved that the constraint of $\mathcal{D}_{\mathbf{x}}(\mathbf{x}_s, \mathbf{\bar{x}}_t)  = \alpha \mathcal{D}_{\mathbf{y}}(\mathbf{y}_s, \mathbf{\bar{y}}_t)$ is also guaranteed by optimizing   Eqn. \eqref{eqn:loss_for_lt}. Indeed, as $\mathcal{D}_{\mathbf{x}}$ and $\mathcal{D}_{\mathbf{y}}$ are defined as distance metrics, for all $\mathbf{x}_t$ and $\mathbf{y}_t = F_{\theta}(\mathbf{x}_t)$, the following triangular inequality hold:
\begin{equation} \label{eqn:triangluar}
\begin{split}
    \mathcal{D}_{\mathbf{x}}(\mathbf{x}_s, \mathbf{x}_t) + \mathcal{D}_{\mathbf{x}}(\mathbf{x}_t, \mathbf{\bar{x}}_t)  &\geq \mathcal{D}_{\mathbf{x}}(\mathbf{x}_s, \mathbf{\bar{x}}_t) \\
    \mathcal{D}_{\mathbf{y}}(\mathbf{y}_s, \mathbf{y}_t) + \mathcal{D}_{\mathbf{y}}(\mathbf{y}_t, \mathbf{\bar{y}}_t)  &\geq \mathcal{D}_{\mathbf{y}}(\mathbf{y}_s, \mathbf{\bar{y}}_t) \\
\end{split}
\end{equation}
Also, as distances $\mathcal{D}_\mathbf{x}$ and $\mathcal{D}_\mathbf{y}$ are bounded under our distribution shift assumption, the constraints of paired data can be further derived as in Eqn.  \eqref{eqn:upper_bound}.
\begin{equation} \label{eqn:upper_bound}
\begin{split}
    &\mathcal{D}_{\mathbf{x}}(\mathbf{x}_s, \mathbf{\bar{x}}_t) - \alpha\mathcal{D}_{\mathbf{y}}(\mathbf{y}_s, \mathbf{\bar{y}}_t) \\
    &\leq \mathcal{D}_{\mathbf{x}}(\mathbf{x}_s, \mathbf{x}_t) + \mathcal{D}_{\mathbf{x}}(\mathbf{x}_t, \mathbf{\bar{x}}_t) - \alpha\mathcal{D}_{\mathbf{y}}(\mathbf{y}_s, \mathbf{\bar{y}}_t)\\
    &\leq \mathcal{D}_{\mathbf{x}}(\mathbf{x}_s, \mathbf{x}_t) + \beta - \alpha(\mathcal{D}_{\mathbf{y}}(\mathbf{y}_s, \mathbf{y}_t)+\beta)+\alpha\beta \\
    &\leq \mathcal{D}_{\mathbf{x}}(\mathbf{x}_s, \mathbf{x}_t)   -\alpha\mathcal{D}_{\mathbf{y}}(\mathbf{y}_s, \mathbf{y}_t) + (1+\alpha)\beta
\end{split}
\end{equation}
In Eqn. \eqref{eqn:upper_bound}, the $\mathcal{D}_{\mathbf{x}}(\mathbf{x}_s, \mathbf{x}_t)   -\alpha\mathcal{D}_{\mathbf{y}}(\mathbf{y}_s, \mathbf{y}_t) + (1+\alpha)\beta$ is an upper bound of $\mathcal{D}_{\mathbf{x}}(\mathbf{x}_s, \mathbf{\bar{x}}_t)  - \alpha \mathcal{D}_{\mathbf{y}}(\mathbf{y}_s, \mathbf{\bar{y}}_t)$. 
In other words, minimizing $||\mathcal{D}_{\mathbf{x}}(\mathbf{x}_s, \mathbf{x}_t) - \alpha\mathcal{D}_{\mathbf{y}}(\mathbf{y}_s, \mathbf{y}_t)||_2^2$ (as $\alpha$ and  $\beta$ are constant numbers, these can be excluded during training) is equivalent to imposing the constraint of $||\mathcal{D}_{\mathbf{x}}(\mathbf{x}_s, \mathbf{\bar{x}}_t)  - \alpha \mathcal{D}_{\mathbf{y}}(\mathbf{y}_s, \mathbf{\bar{y}}_t)||_2^2$. Therefore, the constraints in Eqn. \eqref{eqn:obj_with_cond_pair} is guaranteed when optimizing $\mathcal{L}_t$ defined in Eqn. \eqref{eqn:loss_for_lt}.
Fig. \ref{fig:proposed_framework}(A) illustrates our proposed CROVIA framework.

\noindent
\textbf{Topological Preserving Between Image and Segmentation Spaces}
As shown in Eqn. \eqref{eqn:loss_for_lt}, the loss is defined to optimize the constraint of $||\mathcal{D}_{\mathbf{x}}(\mathbf{x}_s, \mathbf{y}_t) - \alpha\mathcal{D}_{\mathbf{y}}(\mathbf{y}_s, \mathbf{y}_t)||_2^2$ between all unpaired samples. One important property of this constraint is that by enforcing the constraint of Eqn. \eqref{eqn:loss_for_lt}, the cross-view correlations (or relative structures) of images ($\mathcal{D}_{\mathbf{x}}(\mathbf{x}_s, \mathbf{y}_t)$) and segmentation maps ($\mathcal{D}_{\mathbf{y}}(\mathbf{y}_s, \mathbf{y}_t)$) are proportionally equivalent. Therefore, the cross-view topological structures of image distributions is preserved in cross-view segmentation distributions.

\subsection{The Choice of Correlation Metrics}
One of the key factors affecting the performance of our approach is the choice of $\mathcal{D}_{\mathbf{x}}$ and $\mathcal{D}_{\mathbf{y}}$.
The direct metric measured on the image (and segmentation) space, i.e. $\ell_2$, could be adopted in our approach. However, it is ineffective 
because the structural and semantic information is not well captured in the direct metric \cite{duong2020vec2face}.
Therefore, to address this limitation, we proposed to measure $\mathcal{D}_{\mathbf{x}}$ (and $\mathcal{D}_{\mathbf{y}}$) by comparing deep features produced by deep networks where the semantic and structural information of images (and segmentations) is embedded in their deep representations.
Formally, distances $\mathcal{D}_{\mathbf{x}}$ and $\mathcal{D}_{\mathbf{y}}$ can be formulated as:
\begin{equation} \label{eqn:dist_deep}
\begin{split}
    \mathcal{D}_{\mathbf{x}}(\mathbf{x}_s, \mathbf{x}_t) &= \mathcal{D}_{G_\mathbf{x}}(G_{\mathbf{x}}(\mathbf{x}_s), G_{\mathbf{x}}(\mathbf{x}_t)) \\ \mathcal{D}_{\mathbf{y}}(\mathbf{y}_s, \mathbf{y}_t) &= \mathcal{D}_{G_\mathbf{y}}(G_{\mathbf{y}}(\mathbf{y}_s), G_{\mathbf{y}}(\mathbf{y}_t))
\end{split}
\end{equation}
where $G_{\mathbf{x}}$ and $G_{\mathbf{y}}$ are the deep neural networks, $\mathcal{D}_{G_\mathbf{x}}$ and $\mathcal{D}_{G_\mathbf{y}}$ are distances defined in the deep representations.
The distances defined in Eqn. \eqref{eqn:dist_deep}
provide more meaningful measurement as the deep semantic and structural contents of images (or segmentation maps) are embedded in their deep representations.
Several prior works have also adopted this approach \cite{johnson2016perceptual, duong2020vec2face}.
Hence, designing $G_\mathbf{x}$ (and $G_\mathbf{y}$) plays a vital role in our approach.
Directly adopting the common design of CNNs or Transformers could contain some potential limitations as many prior adversarial works \cite{engstrom2019learning, santurkar2019computer, duong2020vec2face} have shown that there could be two different scenes (or segmentation maps) that have similar deep representations but significantly different image content if the deep networks are not bijective.
Then, the distances computed by non-bijective deep networks could not fully reflect the true correlation between two images (and segmentation maps).
To address this problem, we propose to model $G_{\mathbf{x}}$ and $G_{\mathbf{y}}$ (detailed in Sec. \ref{sec:GX_GY_Learn}) as the bijective networks where each image (and segmentation map) is mapped into a unique deep representation in the latent space.

\section{Learning Deep Bijective Networks} \label{sec:GX_GY_Learn}

To effectively learn bijective networks $G_{\mathbf{x}}$ and  $G_{\mathbf{y}}$ so that %
they are able to capture semantic and structural information, 
$G_{\mathbf{x}}$ and  $G_{\mathbf{y}}$ are modeled as the multi-scale structures of bijective networks \cite{dinh2017density, glow, truong2021bimal}
learned by the log-likelihood loss
with the tractable log-determinant computation.

\subsection{Learning Multi-modal Bijective Network on RGB Images}

As $G_\mathbf{x}: \mathcal{X} \to \mathcal{Z}_\mathbf{x}$ is the bijective networks that map an image into the latent space, $G_\mathbf{x}$ can be learned by minimizing the negative log-likelihood with the tractable log-determinant computation \cite{dinh2017density, dinh2017density, truong2021bimal}. 
However, straightforwardly learning $G_\mathbf{x}$
on images of both on-road and UAV views is not optimal.
Indeed, the bijective network is a homomorphism mapping due to its invertible property,  
and therefore it preserves the topological structures of the data domain \cite{DBLP:conf/icml/CornishCDD20, zhang2021diffusion}
In other words, it limits the capability of the bijective network in learning multi-modal data. 

\noindent
\textbf{Global Structure Learning in Image Spaces:} To alleviate this limitation, we propose to disentangle the learning process of the bijective network by conditioning the view (domain) information. 
In particular, 
we proposed a new multi-modal bijective network $G_\mathbf{x}$ that takes an image $\mathbf{x}_s$ and its view information, i.e., $d \in \{s, t\}$, as the network inputs.
Then, learning $G_x$ can be modeled by simultaneously optimizing the negative log-likelihood on both on-road and UAV view images defined as in Eqn. \eqref{eqn:multi_modal_nvp}.
\begin{equation} \label{eqn:multi_modal_nvp}
\footnotesize
\begin{split}
    &\arg\min_{G_\mathbf{x}}-\mathbb{E}_{\mathbf{x} \in \mathcal{X}_s \cup \mathcal{X}_t, d \in \{s, t\}} \log p(\mathbf{x}, d) \\
    = &\arg\min_{G_\mathbf{x}}\Bigg[-\mathbb{E}_{\mathbf{x} \in \mathcal{X}_s}\log p(d=s) \pi(\mathbf{z}_{\mathbf{x}_s}| d=s )\left|\frac{\partial G_\mathbf{x}(\mathbf{x}, s)}{\partial (\mathbf{x}, s)}\right| \\
    &\qquad\qquad\quad -\mathbb{E}_{\mathbf{x} \in \mathcal{X}_t}\log p(d=t) \pi(\mathbf{z}_{\mathbf{x}_t}| d=t)\left|\frac{\partial G_\mathbf{x}(\mathbf{x}, t)}{\partial (\mathbf{x}, t)}\right|\Bigg]
\end{split}
\end{equation}
where $\mathbf{z}_{\mathbf{x}_s} = G_\mathbf{x}(\mathbf{x}, s)$, $\mathbf{z}_{\mathbf{x}_t} = G_\mathbf{x}(\mathbf{x}, t)$, 
$\pi$ is the prior distribution, 
$\left|\frac{\partial G_\mathbf{x}(\mathbf{x}, s)}{\partial (\mathbf{x}, s)}\right|$ is the Jacobian determinant of $G_\mathbf{x}(\mathbf{x}, s)$ w.r.t the input $(\mathbf{x}, s)$ (similar for $\left|\frac{\partial G_\mathbf{x}(\mathbf{x}, t)}{\partial (\mathbf{x}, t)}\right|$),
$p(d=s)$ and $p(d=t)$ are constant numbers that can be statistically computed from the dataset.
The normal distribution has been adopted for the prior distribution $\pi$ in our experiments.

\subsection{Learning Multi-modal Bijective Network on Segmentation Maps}

Learning $G_{\mathbf{y}}: \mathcal{Y} \to \mathcal{Z}_\mathbf{y}$ is not trivial due to the lack of labels in the UAV domain.
Nevertheless, as revealed in Sec \ref{sec:learn_on_unpair}, the relational information between image and segmentation spaces, i.e., the topology of
the samples across views between image and segmentation spaces, are preserved.

\noindent
\textbf{Global Structure Learning in Segmentation Spaces:} Upon the prior topological knowledge, %
$G_\mathbf{y}$ can be learned in a way so that 
the cross-view topology of segmentation distributions generated by $G_\mathbf{y}$ is optimized as close as the cross-view topology in image spaces. 
Hence, along with learning negative log-likelihood on on-road-view segmentation maps, a regularizer $\mathcal{R}$ imposing the topology-preserving is introduced to our learning process as follows:
\begin{equation} \label{eqn:learn_Gy}
\footnotesize
\begin{split}
    \arg\min_{G_\mathbf{y}}\Bigg[-\mathbb{E}_{\mathbf{y} \in \mathcal{Y}_s}\log p(d=s) \pi(\mathbf{z}_{\mathbf{y}_s}| d=s )\left|\frac{\partial G_\mathbf{y}(\mathbf{y}, s)}{\partial (\mathbf{y}, s)}\right| \\
    + \mathbb{E}_{\mathbf{z}_{\mathbf{y}_t} \sim \pi(\mathbf{z}_{\mathbf{y}_t} | d=t)}\mathcal{R}(\widetilde{\mathbf{y}}_t)\Bigg]
\end{split}
\end{equation}
where $\mathcal{Y}_s$ is the set of on-road-view segmentation maps,
$\mathbf{z}_{\mathbf{y}_s} = G_\mathbf{y}(\mathbf{y}, s)$, $\widetilde{\mathbf{y}}_t = G_\mathbf{y}^{-1}(\mathbf{z}_{\mathbf{y}_t}, t)$, 
$\left|\frac{\partial G_\mathbf{y}(\mathbf{y}, s)}{\partial (\mathbf{y}, s)}\right|$ is the Jacobian determinant of $G_\mathbf{y}(\mathbf{y}, s)$ w.r.t the input $(\mathbf{y}, s)$. 
In Eqn. \eqref{eqn:learn_Gy}, the first term aims to capture deep semantic and structural information of segmentation maps learned on the on-road-view data.  
Meanwhile, the second term aims to enforce the topology of generated target segmentation distributions. Thus, $\mathbb{E}_{\mathbf{z}_{\mathbf{y}_t} \sim \pi(\mathbf{z}_{\mathbf{y}_t} | d=t)}\mathcal{R}(\widetilde{\mathbf{y}}_t)$  can be defined as, %
\begin{equation}\label{eqn:GW_distance_Gy}
\begin{split}
    \min_{\sigma}\mathbb{E}_{\mathbf{x}_s, \mathbf{x}_t, \mathbf{y}_s, \widetilde{\mathbf{y}}_t}||\ell_2(\mathbf{x}_s, \mathbf{x}_t) - 
    \alpha\ell_2(\mathbf{y}_{s}, \widetilde{\mathbf{y}}_t)||_2^2\sigma_{\mathbf{x}_s, \mathbf{y}_s}\sigma_{\mathbf{x}_t, \widetilde{\mathbf{y}}_t}
\end{split}
\end{equation}
where $\sigma$ is an association matrix. 
As the corresponding pair of image $\mathbf{x}_t$ and synthesized segmentation $\widetilde{\mathbf{y}}_t$ is unknown during training, the association matrix $\sigma$ is introduced to approximate the corresponding pair between them
where $\sigma_{\mathbf{x}_t, \widetilde{\mathbf{y}}_t}$ denotes the probability of association between $\mathbf{x}_t$ and $\widetilde{\mathbf{y}}_t$. 
As $\mathbf{y}_s$ is the segmentation of image $\mathbf{x}_s$, the association between them can be set to $1$, i.e., $\sigma_{\mathbf{x}_s, \mathbf{y}_s}=1$.
Eqn. \eqref{eqn:GW_distance_Gy} can be solved by the Gromov-Wasserstein solver \cite{DBLP:conf/icml/BunneA0J19, vay_sgw_2019}. We adopt 
the learning approach of \cite{DBLP:conf/icml/BunneA0J19, vay_sgw_2019} to optimize Eqn. \eqref{eqn:learn_Gy}. 
Fig. \ref{fig:proposed_framework}(B) illustrates our learning framework of the multi-model bijective network.

\noindent
\textbf{Correlation Metrics: }
Finally, distances $\mathcal{D}_\mathbf{x}$ and $\mathcal{D}_\mathbf{y}$ measured via $G_\mathbf{x}$ and $G_\mathbf{y}$ 
can be defined as the squared Wasserstein coupling distance between two Gaussian distributions:
\begin{equation}
\small
\begin{split}
    &\mathcal{D}_{\mathbf{x}}(\mathbf{x}_s, \mathbf{x}_t) 
    = \mathcal{D}_{G_\mathbf{x}}(G_{\mathbf{x}}(\mathbf{x}_s, s), G_{\mathbf{x}}(\mathbf{x}_t), t) = \inf \mathbb{E}(||\mathbf{z}_{\mathbf{x}_s}-\mathbf{z}_{\mathbf{x}_t}||_2^2)\\
    &= || \mu_{\mathbf{x}_s}-\mu_{\mathbf{x}_t} ||^2_2 
    + \text{Tr}(\Sigma_{\mathbf{x}_s} + \Sigma_{\mathbf{x}_t} - 2(\Sigma_{\mathbf{x}_s}^{1/2}\Sigma_{\mathbf{x}_t}\Sigma_{\mathbf{x}_s}^{1/2})^{1/2})
\end{split}
\end{equation}
where $\{\mu_{\mathbf{x}_s}, \Sigma_{\mathbf{x}_s}\}$ and $\{\mu_{\mathbf{x}_t}, \Sigma_{\mathbf{x}_t}\}$ are the means and covariances of $\mathbf{z}_{\mathbf{x}_s}$ and $\mathbf{z}_{\mathbf{x}_t}$.
To satisfy the bounded distribution shift assumption, 
the distance $\mathcal{D}_\mathbf{x}(\mathbf{x}_s, \mathbf{x}_t)$ is defined as the minimum between its value and bounded value $\beta$, i.e. $\operatorname{min}(\mathcal{D}_\mathbf{x}(\mathbf{x}_s, \mathbf{x}_t), \beta)$.
The distance $\mathcal{D}_\mathbf{y}$ 
is also defined as similar to $\mathcal{D}_\mathbf{x}$.
The $\beta$ value is set to $100$ in our experiments.

\section{Experiments}

In this section, we first review datasets, implementation, and evaluation benchmarks. Then, the ablation studies analyze the effectiveness of our proposed approach. Finally, we compare our SOTA results with prior UDA approaches.

\subsection{Datasets, Implementations, and Benchmarks}

\noindent
\textbf{UAVID} \cite{uavid_dataset} is a real-world UAV scene segmentation dataset. This dataset collected in the urban streets includes 42 video sequences at the 4K resolution %
in slanted views. Segmentation of UAVID is challenging due to the high spatial resolution, spatial variations, and complex scenes.

\noindent
\textbf{SYNTHIA} \cite{Ros_2016_CVPR} is a synthetic segmentation dataset including $9,400$ pixel-level labelled RGB images. It is generated from a simulator in scene segmentation of urban settings. 

\noindent
\textbf{GTA5} \cite{Richter_2016_ECCV} is a synthetic dataset created from the game engine. This dataset includes  $24,966$ synthetic, densely labeled images with 33 class categories at high resolution.

\noindent
\textbf{Implementation}
In our experiments, two different segmentation network architectures are used, i.e.,
DeepLab-V2 \cite{chen2018deeplab} network with a ResNet-101 \cite{he2015deep} backbone and Transformer \cite{daformer} with a MiT-B4 encoder \cite{xie2021segformer}.
Following the UAV protocol of \cite{UNetFormer}, the image size is set to $1024 \times 1024$. 
The design of $G_{\mathbf{x}}$ and $G_{\mathbf{y}}$ is identical. 
In particular, our multi-model bijective network is designed as multi-scale architecture adopted from \cite{alignflow} where each scale includes multiple steps of the flow.
Every single flow step injected by the domain information is designed as a stack of AcNorm, Invertible $1 \times 1$ Convolution, and Residual-style Affine Coupling Layer \cite{dinh2017density, glow, truong2021bimal}. 
The number of scales and flows in our experiments are set to $4$ and $32$, respectively. 
The entire framework is optimized by the SGD optimizer on four 48GB-VRAM GPUs, where the batch size of each GPU is set to $8$ and the base learning rate is set to $2.5\times10^{-4}$. 
To increase the diversity of training data, several data augmentation techniques \cite{Araslanov:2021:DASAC, shi2020where} 
are adopted in the training process.

\begin{table}[b]
\centering
\small
\caption{Semantic Segmentation  mIoU Performance (\%) using DeepLab-V2 on a validation set of UAVID 
w.r.t. Different Values of $\alpha$.}
\label{tab:alpha_ab}
\setlength{\tabcolsep}{3pt}
\begin{tabular}{c | c c c c c c | c}
\hline
$\alpha$     &  Road & Build. & Car  & Tree & Terrain & Person& mIoU \\ 
\hline
\multicolumn{8}{c}{SYNTHIA $\to$ UAVID} \\ \hline
0.1          & 6.7           & 63.7              & 47.1          & 51.0          & $-$             & 14.9                & 36.7          \\ %
0.5          & 7.0           & 64.0              & 47.2          & 51.1          & $-$             & 15.1                & 36.9          \\ %
1.0          & 7.3           & 64.1              & 47.5          & 51.4          & $-$             & 15.5                & 37.2          \\ %
1.5          & 7.4           & 64.2              & 48.3          & 51.7          & $-$             & 15.8                & 37.5          \\ %
\textbf{2.0} & \textbf{10.6} & \textbf{65.7}     & \textbf{51.7} & \textbf{55.6} & \textbf{$-$}    & \textbf{17.0}       & \textbf{40.1} \\ %
2.5          & 9.4           & 65.3              & 50.8          & 54.3          & $-$             & 16.7                & 39.3          \\ %
3.0          & 8.2           & 64.7              & 49.7          & 51.7          & $-$             & 16.6                & 38.2          \\ \hline
\multicolumn{8}{c}{GTA5 $\to$ UAVID} \\ \hline
0.1          & 4.7           & 44.6              & 7.9           & 37.3          & 39.4            & 4.9                 & 23.1          \\ %
0.5          & 5.4           & 44.6              & 8.0           & 38.9          & 39.5            & 5.1                 & 23.6          \\ %
1.0          & 6.4           & 45.4              & 8.2           & 41.4          & 40.0            & 6.1                 & 24.6          \\ %
1.5          & 9.7           & 46.5              & 8.3           & 41.9          & 41.5            & 6.5                 & 25.7     \\ 
2.0          & 14.7          & 49.3              & 9.4           & 47.9          & 42.9            & 7.8                 & 28.7          \\ %
\textbf{2.5} & \textbf{18.2} & \textbf{49.8}     & \textbf{10.4} & \textbf{48.1} & \textbf{44.0}   & \textbf{8.0}        & \textbf{29.7} \\ %
3.0          & 14.9          & 49.3              & 10.1          & 47.9          & 43.4            & 7.8                 & 28.9          \\ %
\hline
\end{tabular}
\end{table}

\begin{figure}[t]
    \centering
    \includegraphics[width=0.48\textwidth]{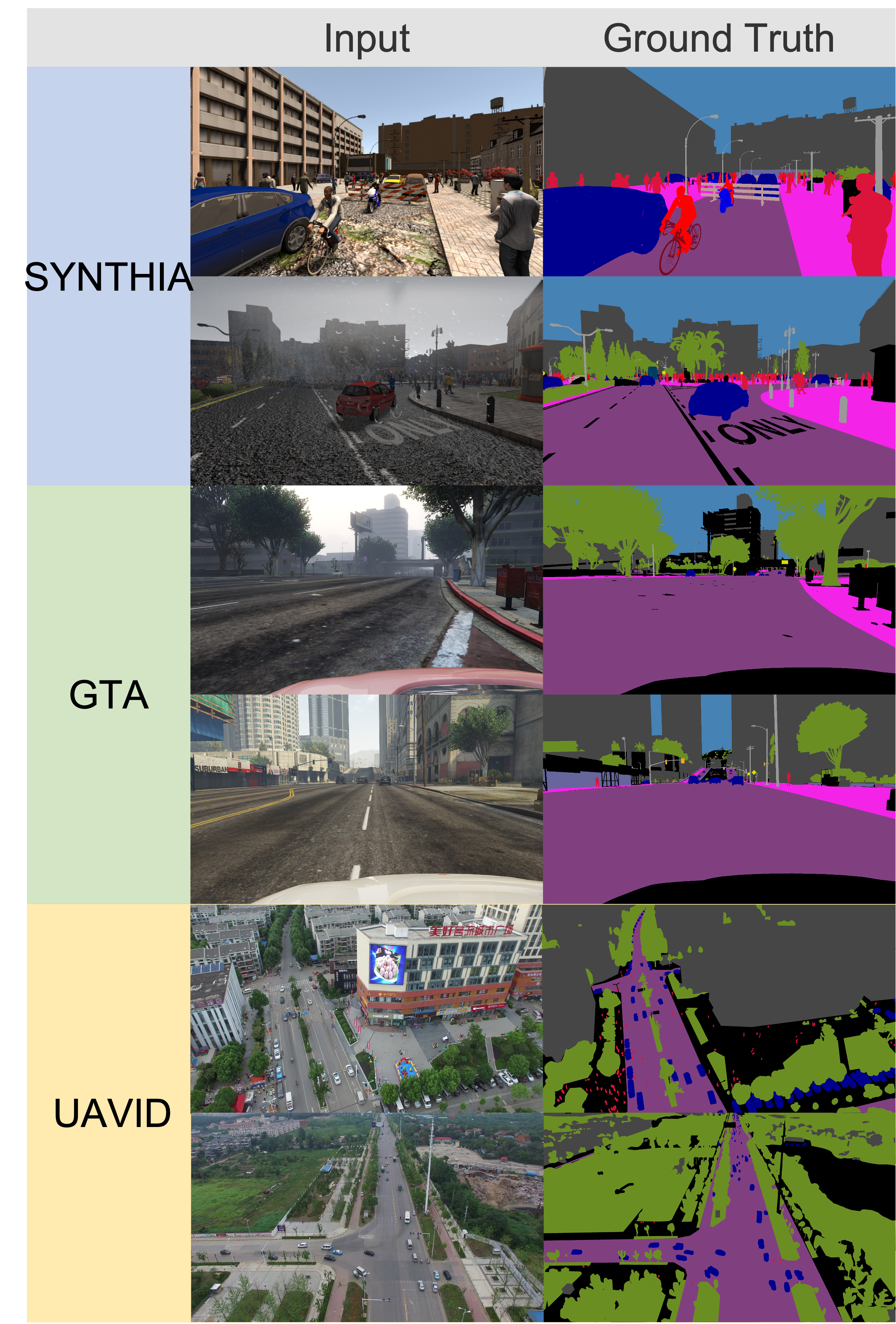}
    \caption{The Input and Ground-Truth Samples of SYNTHIA, GTA5, and UAV Datasets.}
    \label{fig:dataset_sample}
\end{figure}

\noindent
\textbf{Benchmark Protocols}

We present two new benchmarks for the cross-view adaptation task based on the SYNTHIA, GTA5, and UAVID datasets. Figure \ref{fig:dataset_sample} illustrates the input and ground-truth samples of these datasets.
Motivated by UAD research \cite{vu2019advent, vu2019dada, daformer, Araslanov:2021:DASAC}, SYNTHIA and GTA5 have been widely adopted as their standard benchmarks. 
SYNTHIA and GTA5 are chosen as: (1) they have a great overlap class of interests with UAVID, and 
help to efficiently illustrate and evaluate the adaptation task and (2) these are widely used in UDA benchmarks \cite{vu2019advent, vu2019dada, daformer, Araslanov:2021:DASAC}.
In the original UAVID dataset, there are eight categories, i.e. Building, Road,  Static Car,  Moving Car, Tree, Terrain,  Person, and  Background Clutter.
However, in SYNTHIA and GTA5, the moving cars and static cars are not distinguished. Thus, we consider these two classes a single class ( \textit{`Car'}).
Besides, in the UAVID dataset, Cars, Trucks, and Buses are all annotated as a class of Car. 
Hence, these three classes in SYNTHIA and GTA5 datasets is also considered a single class (\textit{`Car'}).
Also, following the prior UDA benchmarks (i.e., SYNTHIA $\to$ Cityscapes and GTA5 $\to$ Cityscapes \cite{vu2019advent, vu2019dada, truong2021bimal, zhang2021prototypical, Araslanov:2021:DASAC, daformer}), the class of background clutter is excluded.
Therefore, we introduce two new benchmarks for cross-view adaptation, i.e., SYNTHIA $\to$ UAVID and GTA5 $\to$ UAVID. 
In the SYNTHIA $\to$ UAVID benchmark, as the class of Terrain is not available  SYNTHIA, this class is excluded. 
In summary, five classes in common between SYNTHIA and UAVID are selected, i.e. Road, Building, Car, Tree, and Person.
In the GTA5 $\to$ UAVID benchmark, it has six classes in common including five classes of the SYNTHIA $\to$ UAVID benchmark and one more class of Terrain.
In experiments, the performance of segmentation models is measured by the mean Intersection over Union (mIoU) metric.

\subsection{Ablation Study}

\noindent
\textbf{Effectiveness of Choosing $\alpha$:}
To illustrate the effectiveness of value $\alpha$ in Eqn. \eqref{eqn:loss_for_lt}, we evaluate our models using DeepLab-V2 with different values of $\alpha$ ranging from $0.1$ to $3.0$ on two benchmarks, i.e. SYNTHIA $\to$ UAVID and GTA5 $\to$ UAVID.
Distances $\mathcal{D}_{\mathbf{x}}$ and $\mathcal{D}_{\mathbf{y}}$ computed via $G_\mathbf{x}$ and $G_\mathbf{y}$ are utilized in this experiment.
As shown in Table \ref{tab:alpha_ab}, the mIoU performance is consistently improved w.r.t the increase of value $\alpha$ from $0.1$ to $2.0$ on  SYNTHIA $\to$ UAVID and from $0.1$ to $2.5$ on  GTA5 $\to$ UAVID. 
The optimal value $\alpha$ on SYNTHIA $\to$ UAVID and GTA5 $\to$ UAVID benchmarks are $2.0$ and $2.5$ where 
our mIoU accuracy achieves $40.1\%$ and $29.7\%$, respectively.
Later, the mIoU performance tends to steadily drop when $\alpha$ increases. 
Basically, the variation in the image space is typically higher than in the segmentation space due to the higher complexity of image data,
i.e., images have a more complex shape, object textures, and appearance; meanwhile, the segmentation maps represent objects based on their categories with less complexity in textures and appearance. Then, if $\alpha$ is small, it could not represent the correct proportion of changes between images and segmentation maps.
Meanwhile, the higher value of $\alpha$ tends to exaggerate the changes in segmentation maps leading to the performance drop when $\alpha$ keeps increasing over the optimal value.

\begin{table}[!t]
\centering
\small
\caption{Semantic Segmentation mIoU Performance (\%) on the validation set of UAVID
Using DeepLab V2 and Transformer (Trans.) w.r.t the choice of $\mathcal{D}_{\mathbf{x}}$ and $\mathcal{D}_{\mathbf{y}}$, i.e.  (A) Without Cross-View Adaptation, (B) Direct Distances ($\ell_2$), (C) Distances computed by the Pure Bijective Networks, and (D) Distances computed by the Multi-modal Bijective Networks.
}
\label{tab:distance_ab}
\setlength{\tabcolsep}{3pt}
\begin{tabular}{c c | c c c c c c |c}
\hline
\multicolumn{2}{c|}{\textbf{Config}}              & 
Road & Build. & Car  & Tree & Terrain & Person& mIoU \\ 
\hline 
\multicolumn{9}{c}{SYNTHIA $\to$ UAVID} \\
\hline 
\multirow{4}{*}{\begin{tabular}{@{}c@{}} DeepLab\\V2  \end{tabular}} & (A)  & 3.7           & 59.5              & 36.8          & 32.4          & $-$             & 7.9                 & 28.1 \\

& (B)         & 4.2           & 62.4          & 37.6          & 49.9          & $-$           & 11.8          & 33.2          \\
                            & (C)           & 7.0           & 63.9          & 47.1          & 51.0          & $-$           & 15.1          & 36.8          \\
                            & \textbf{(D)} & \textbf{10.6} & \textbf{65.7} & \textbf{51.7} & \textbf{55.6} & \textbf{$-$}  & \textbf{17.0} & \textbf{40.1} \\
\hline
\multirow{4}{*}{Trans.}   & (A)  & 5.6 &	58.8 &	36.0 &	50.9 &	$-$	& 10.1	& 32.3\\
& (B)         & 6.3           & 63.7          & 46.7          & 50.9          & $-$           & 13.7          & 36.2          \\
                            & (C)           & 12.6          & 69.2          & 52.8          & 57.3          & $-$           & 17.4          & 41.9          \\
                            & \textbf{(D)}          & \textbf{16.3} & \textbf{75.1} & \textbf{59.6} & \textbf{60.0} & $-$           & \textbf{19.1} & \textbf{46.0} \\
\hline
\multicolumn{9}{c}{GTA5 $\to$ UAVID} \\
\hline
\multirow{4}{*}{\begin{tabular}{@{}c@{}} DeepLab\\V2  \end{tabular}} & (A)  & 2.1           & 49.8              & 6.8           & 21.0          & 22.2            & 0.0                 & 17.0          \\ 

& (B)         & 3.7           & 43.6          & 7.0           & 33.9          & 37.7          & 3.7           & 21.6          \\
                            & (C)           & 14.8          & 47.7          & 28.7          & 18.8          & 36.8          & 6.3           & 25.5          \\
                            & \textbf{(D)} & \textbf{18.2} & \textbf{49.8} & \textbf{10.4} & \textbf{48.1} & \textbf{44.0} & \textbf{8.0}  & \textbf{29.7} \\
\hline
\multirow{4}{*}{Trans.}     & (A) & 3.0	& 37.0	& 7.9	& 41.4	& 43.4	& 8.3	& 23.5 \\
& (B)         & 16.6          & 53.0          & 27.2          & 24.4          & 36.6          & 4.5           & 27.0          \\
                            & (C)           & 19.9          & 53.6          & 14.9          & 50.7          & 45.3          & 10.9          & 32.6          \\
                            & \textbf{(D)} & \textbf{20.5} & \textbf{56.1} & \textbf{37.6} & \textbf{50.7} & \textbf{45.3} & \textbf{10.9} & \textbf{36.8} \\
\hline
\end{tabular}
\end{table}

\noindent
\textbf{Effectiveness of Distances $\mathcal{D}_\mathbf{x}$ and $\mathcal{D}_\mathbf{y}$:} 
The models are evaluated with  different settings of $\mathcal{D}_\mathbf{x}$ and $\mathcal{D}_\mathbf{y}$ on two benchmarks.
The value of $\alpha$ is set by the optimal value in previous experiments.
There are four settings evaluated, 
i.e. \textbf{(A)} The model is trained on the source dataset only,
\textbf{(B)} Distances $\mathcal{D}_{\mathbf{x}}$ and $\mathcal{D}_{\mathbf{y}}$ are computed by the direct metric (i.e., $\ell_2$), 
\textbf{(C)} Distances $\mathcal{D}_{\mathbf{x}}$ and $\mathcal{D}_{\mathbf{y}}$ are computed via the pure bijective networks without domain conditions,
\textbf{(D)} Distances $\mathcal{D}_{\mathbf{x}}$ and $\mathcal{D}_{\mathbf{y}}$ are computed via our multi-modal bijective networks. 
As results in Table \ref{tab:distance_ab}, 
our mIoU result of config (B) outperforms config (A), i.e., our CROVIA (Transformer) using $\ell_2$ as correlation metrics has improved mIoU accuracy to $36.2\%$ and $27.0\%$ on SYNTHIA $\to$ UAVID and GTA5 $\to$ UAVID benchmarks.
With configs (C)-(D) where $\mathcal{D}_{\mathbf{x}}$ and $\mathcal{D}_{\mathbf{y}}$  are computed by deep networks, the mIoU results are significantly boosted, i.e., our CROVIA (Transformer) with config (D) has achieved the SOTA results on SYNTHIA $\to$ UAVID and GTA5 $\to$ UAVID, which are $46.0\%$ and $37.8\%$, respectively.
These improvements are totally explainable because the direct distance $\ell_2$ is quite sensitive to the changes of each pixel and the semantic information and global structures of images are not well captured by $\ell_2$.
Meanwhile, by using deep networks as configs (C)-(D), the distance metrics provide more meaningful measurements as the semantic and structural contents of images (segmentation maps) are well embedded in their deep representations. 

\noindent
\textbf{Effectiveness of Multi-modal Bijective Networks}
Experimental results in Table \ref{tab:distance_ab} have shown the de facto role of our multi-modal bijective networks.
Because of the homomorphic property of pure bijective networks in config (C), it has limited the capability of $G_\mathbf{x}$ and $G_\mathbf{y}$ in modeling multi-modal data.
Meanwhile, with the awareness of the domain condition, our proposed multi-modal bijective networks in config (D) are able to disentangle the multi-modal data and increase the capability of modeling multi-modal data of $G_\mathbf{x}$ and $G_\mathbf{y}$.
Experimental results in Table \ref{tab:distance_ab} have supported our claim.  
Particularly, in Table \ref{tab:distance_ab}, the results of config (D) using our proposed multi-modal bijective network outperform config (C), i.e., the mIoU results of our CROVIA (Transformer) approach on SYNTHIA $\to$ UAVID and GTA5 $\to$ UAVID are improved by $+4.1\%$ and $4.2\%$, respectively.

\noindent
\textbf{Comparison with Supervised Result}
Table \ref{tab:sota_compare}  reports a comparison of our method with a supervised baseline. On the SYNTHIA $\to$ UAVID benchmark, while the supervised result (66.6\%) is an upper-bound result of unsupervised methods, 
our unsupervised result achieves \textbf{40.1\%} which is \textbf{60.21\%} ($=\frac{40.1}{66.6}$) performance of the supervised one.
To better understand the significance and position of our results, 
considering an example of a standard UDA benchmark (SYNTHIA$\to$Cityscapes),
the unsupervised result of AdvEnt \cite{vu2019advent}, an advanced domain adaptation baseline, is 41.2\% and  achieves 57.70\% ($=\frac{41.2}{71.4}$) of the supervised baseline  which is 71.4\% as reported in \cite{chen2018deeplab}. Our empirical results show CROVIA effectively improves performance on the cross-view adaptation task and outperform prior UDA approaches.

\subsection{Comparison with SOTA Methods} \label{sec:quantitative_result}

\begin{table}[t]
\centering
\small
\caption{Comparison of Semantic Segmentation mIoU Performance (\%) with prior UDA methods on the validation set of UAVID 
}
\label{tab:sota_compare}
\setlength{\tabcolsep}{2pt}
\begin{tabular}{l | c c c c c c | c}
\hline
& Road & Build. & Car  & Tree & Terrain & Person& mIoU \\  

\hline
\multicolumn{8}{c}{SYNTHIA $\to$ UAVID} \\ 
\hline

W/O Adapt.              & 3.7           & 59.5              & 36.8          & 32.4          & $-$             & 7.9                 & 28.1          \\ %

AdvEnt \cite{vu2019advent}                   & 4.7           & 63.2              & 31.7          & 48.6          & $-$             & 11.4                & 31.9          \\ %

DADA \cite{vu2019dada}                     & 10.7          & 63.1              & 32.9          & 50.0          & $-$             & 16.2                & 34.6          \\ %

BiMaL \cite{truong2021bimal}                   & 5.4           & 62.1              & 34.8          & 50.7          & $-$             & 12.7                & 33.1          \\ %

SceneAdapt \cite{SceneAdapt} & 5.3  & 62.5 & 28.4 &  48.3 & $-$ & 13.4 &   31.6 \\ 

SAC \cite{Araslanov:2021:DASAC}                     & 13.9          & 64.0              & 18.7          & 48.0          & $-$             & 15.6                & 32.0          \\ %

ProDA \cite{zhang2021prototypical}                   & 10.6          & 64.7              & 34.1          & 44.5          & $-$             & 17.0                & 34.2          \\ %

DAFormer \cite{daformer}                 & 7.3           & 75.1              & 51.7          & 48.0          & $-$             & 15.1                & 39.4          \\ 

\hline
CROVIA - ResNet               & 10.6          & 65.7              & 51.7          & 55.6          & $-$             & 17.0                & 40.1          \\ %

\textbf{CROVIA - Trans.} & \textbf{16.3} & \textbf{75.1}     & \textbf{59.6} & \textbf{60.0} & \textbf{$-$}    & \textbf{19.1}       & \textbf{46.0} \\ 
\hline

Supervised - ResNet & 72.2 & 86.8 & 74.2 & 76.1 & $-$ & 23.9 & 66.6 \\
\hline

\multicolumn{8}{c}{GTA5 $\to$ UAVID} \\ \hline

W/O Adapt.              & 2.1           & 49.8              & 6.8           & 21.0          & 22.2            & 0.0                 & 17.0          \\ %

AdvEnt \cite{vu2019advent}                   & 2.0           & 30.3              & 14.9          & 29.8          & 41.5            & 1.8                 & 20.0          \\ %

BiMaL \cite{truong2021bimal}                    & 1.3           & 44.6              & 10.1          & 49.2          & 20.0            & 10.9                & 22.7          \\ %

 SceneAdapt \cite{SceneAdapt} & 9.6 & 39.9 & 8.2 & 32.1 & 26.2 & 1.3                 & 19.6          \\
 
SAC \cite{Araslanov:2021:DASAC}                      & 4.5           & 36.9              & 7.8           & 47.9          & 44.1            & 7.8                 & 24.8          \\ %

ProDA \cite{zhang2021prototypical}                    & 6.9           & 50.6              & 28.4          & 25.5          & 38.7            & 4.5                 & 25.8          \\ %

DAFormer \cite{daformer}                 & 15.3          & 51.6              & 33.6          & 27.8          & 38.5            & 4.0                 & 28.5          \\ %

\hline
CROVIA - ResNet               & 18.2          & 49.8              & 10.4          & 48.1          & 44.0            & 8.0                 & 29.7          \\ %

\textbf{CROVIA - Trans.} & \textbf{20.5} & \textbf{56.1}     & \textbf{37.6} & \textbf{50.7} & \textbf{45.3}   & \textbf{10.9}       & \textbf{36.8} \\ 
\hline

{Supervised - ResNet} & 72.2 & 86.8 & 74.2 & 76.1 & 67.8 & 23.9 & 66.8 \\
\hline
\end{tabular}
\end{table}

\noindent
\textbf{SYNTHIA $\to$ UAVID:} 
Table \ref{tab:sota_compare} presents the results of our CROVIA approach compared to UDA methods using DeepLab-V2 and Transformer.
In experiments using DeepLab-V2, we compare our results with AdvEnt \cite{vu2019advent}, DADA \cite{vu2019dada}, BiMaL \cite{truong2021bimal}, SAC \cite{Araslanov:2021:DASAC}, and ProDA \cite{zhang2021prototypical}.
Meanwhile, in experiments using Transformer, CROVIA is compared with DAFormer \cite{daformer}.
Our results in Table \ref{tab:sota_compare} have gained the SOTA performance and are higher than prior UDA methods by a large margin.
Particularly, our mIoU accuracy using Transformer has achieved $46.0\%$ which is higher than DAFormer \cite{daformer} by $+6.6\%$. 
Considering per-class results, our method notably promotes the mIoU result of each individual class,
i.e, \textit{`Road'} ($16.3\%$), \textit{`Building'} ($75.1\%$), \textit{`Car'} ($59.6\%$), \textit{`Tree'} ($60.0\%$), and \textit{`Person'} ($19.1\%$). 
These results have shown our CROVIA gaining advanced results in each class compared to prior methods. 
Moreover, it should be noted that by our formulation, our proposed GeiCo loss has inherited the ``geometry aware'' property from Remarks 1-2.
Therefore, during learning, the requirements of depths or camera poses are not necessary since the ``geometry awareness'' is implicitly learned in our GeiCo loss.
 In addition, although CROVIA does not utilize depths, our results outperform DADA \cite{vu2019dada} which utilizes depth labels. This has further confirmed the effectiveness of our proposed method in term of learning geometry awareness.

\begin{figure*}[t]
    \centering
    \includegraphics[width=1.0\textwidth]{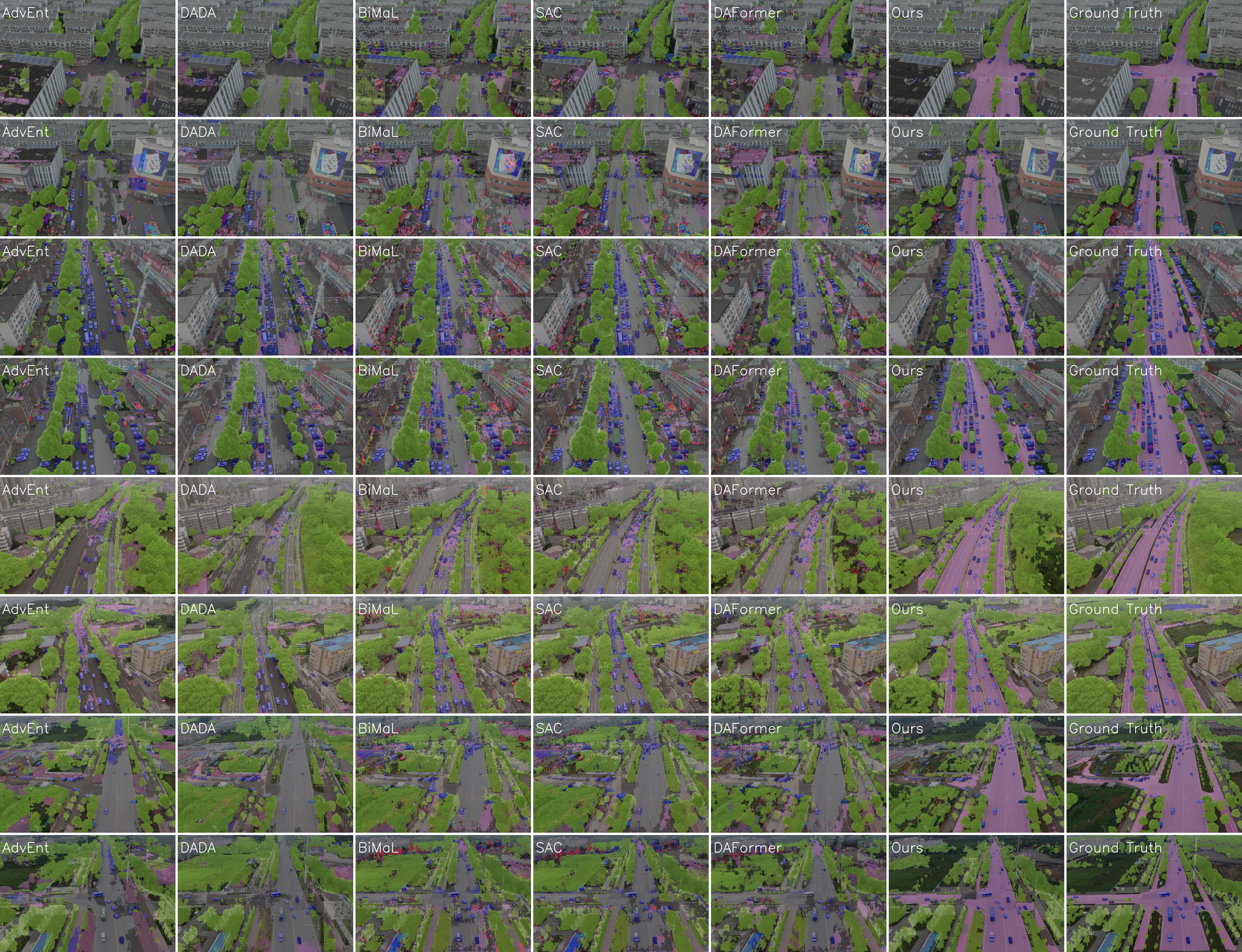}
    \caption{\textbf{Qualitative Results on SYNTHIA $\to$ UAVID}. Columns 1-7 are results of AdvEnt \cite{vu2019advent}, DADA \cite{vu2019dada}, BiMaL \cite{truong2021bimal}, SAC \cite{Araslanov:2021:DASAC}, DAFormer \cite{daformer}, our CROVIA, and Ground Truths (Best view in color and $2\times$ zoom).}
    \label{fig:syn2uavid_qual_res}
\end{figure*}

\noindent
\textbf{GTA5 $\to$ UAVID:}
Experimental results in Table \ref{tab:sota_compare} have shown our CROVIA approach achieving the SOTA performance over six classes of the benchmark GTA5 $\to$ UAVID.
Our CROVIA is compared with prior UDA methods as similar in the SYNTHIA $\to$ UAVID experiment. However, we exclude the DADA \cite{vu2019dada} approach because the GTA5 dataset does not contain depth labels as required by \cite{vu2019dada}.
In this experiment, our proposed methods using DeepLab-V2 and Transformer backbones achieve state-of-the-art performance compared to other approaches using the same backbones.
Specifically, our results using the Transformer backbone achieve the mIoU accuracy of $36.8\%$, higher than DAFormer \cite{daformer} by $+8.3\%$.
Analyzing the mIoU result of each class, in comparison with DAFormer \cite{daformer}, the result of each class is all improved by a large margin with at least $+4.0\%$, specifically, \textit{`road'} ($+5.2\%$), \textit{`building'} ($+4.5\%$), \textit{`car'} ($+4.0\%$), \textit{`tree'} ($+22.9\%$), \textit{`terrain'} ($+6.8\%$)and \textit{`person'} ($+6.9\%$).
These advanced experimental results have shown the effectiveness property of our proposed approach in performing the cross-view adaptation task compared to standard domain adaptation methods.

\noindent
\textbf{Qualitative Results}
Fig. \ref{fig:syn2uavid_qual_res} illustrates the results of our method using the Transformer backbone compared to 
 AdvEnt \cite{vu2019advent}, DADA \cite{vu2019dada},
BiMaL \cite{truong2021bimal}, SAC, \cite{Araslanov:2021:DASAC}, and DAFormer \cite{daformer} on the SYNTHIA $\to$ UAVID benchmark.
Our results have produced better qualitative results compared to prior adaptation methods using adversarial learning adversarial approaches (AdvEnt, DADA, BiMaL) and pseudo-labels (SAC, DAFormer).
In particular, our model is able to accurately identify the border regions of classes, especially in the class of \textit{`road'}.  
The continuity of each object is better than prior methods and matches the ground truth labels.
Although our results are better than other methods, there are a few regions that remain unclear and confusing. For example, the segmentation of the road is discontinued in some regions, and its  boundary with buildings and trees is unsharp and unclear. More qualitative results are available in the supplementary.

\section{Conclusions}

This paper has presented a novel approach for cross-view adaptation in semantic scene segmentation. This paper has sufficiently solved the limitations of the standard domain adaptation by 
introducing the geometric constraint and topological structural constraints into our proposed  Geometry-Constraint Cross-View loss.
Moreover, the proposed multi-modal bijective networks guarantee the proposed GeiCo loss to be able to model global and local structures of the segmentation maps across views. The experiments on two  benchmarks, i.e., SYNTHIA $\to$ UAVID, GTA $\to$ UAVID, have shown the notable performance of our proposed CROVIA approach. Particularly, our CROVIA method achieves the SOTA performance in both benchmarks and improves the performance of the segmentation compared to the prior domain adaptation methods.

\section{Limitations}

\noindent
\textbf{Limitation of hyper-parameter $\alpha$:} Considering the value of $\alpha$ as a constant number in the constraint of geometric correlations across views could bring some potential limitations as the cross-view distances of images and segmentation maps could be a non-linear proportion and may be scaled w.r.t an individual image and its segmentation map.
Future works should consider $\alpha$ as a learned parameter or model $\alpha$ by a deep network to gain more improvement.

\noindent
\textbf{Limitation of bounded value $\beta$:}
Although constraining distances $\mathcal{D}_\mathbf{x}$ and $\mathcal{D}_\mathbf{y}$ bounded by a value $\beta$ under our distribution shift assumption allows us to form the upper bound property as in Eqn. \eqref{eqn:upper_bound} and our GeiCo loss in Eqn. \eqref{eqn:loss_for_lt},
it could contain some potential limitations.
If the changes across views are significantly large, the distances between images (or segmentation maps) could be over bounded value $\beta$. 
Thus, optimizing Eqn. \eqref{eqn:loss_for_lt} could be challenging. The gradients of Eqn. \eqref{eqn:loss_for_lt} will not be differentiable when the value of the distance is greater than $\beta$.

\bibliographystyle{IEEEtran}
\bibliography{references}

\begin{IEEEbiography}[{\includegraphics[width=1in,height=1.25in,clip,keepaspectratio]{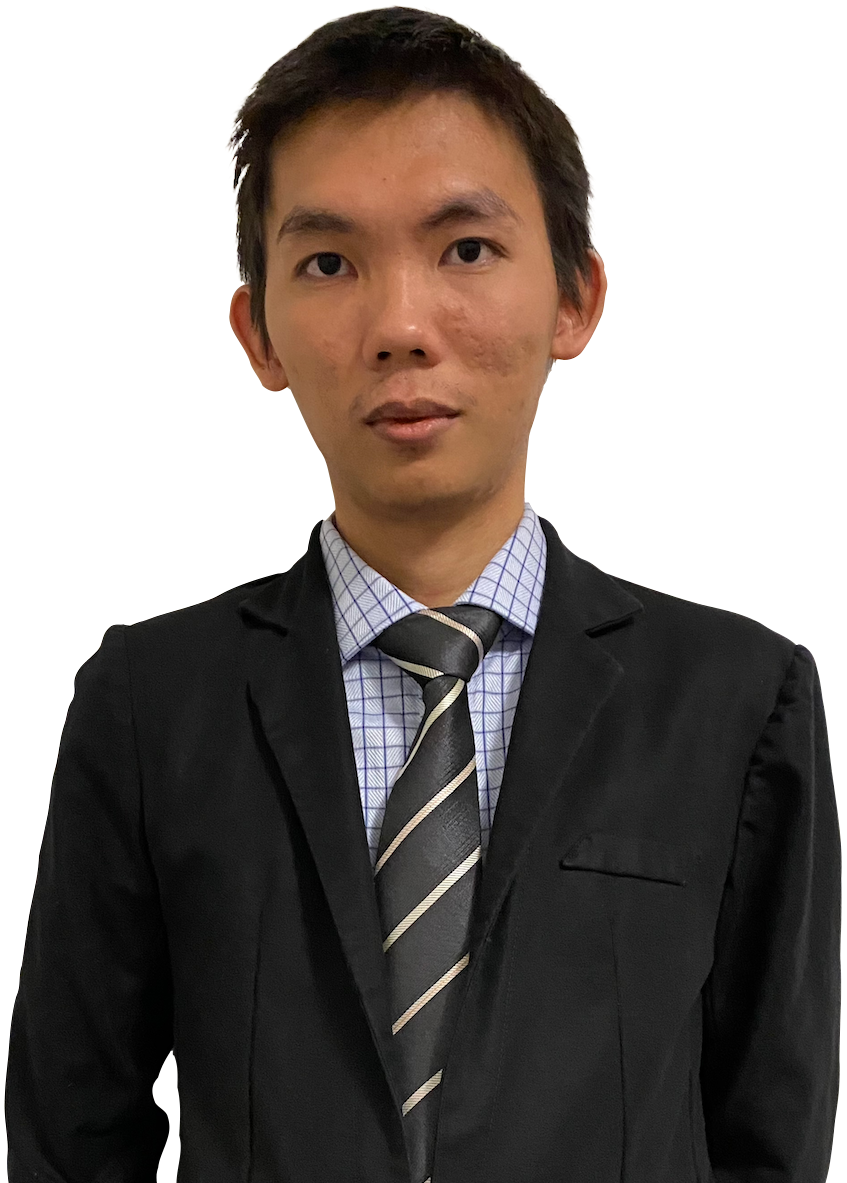}}]{Thanh-Dat Truong}
is currently a Ph.D. Candidate at the Department of Computer Science and Computer Engineering of the University of Arkansas. He received his B.Sc. degree in Computer Science from Honors Program, University of Science, VNU in 2019. He was a research intern at Coordinated Lab Science at the University of Illinois at Urbana-Champaign in 2018. When Thanh-Dat Truong was an undergraduate student, he worked as a research assistant at Artificial Intelligence Lab at the University of Science, VNU. Thanh-Dat Truong's research interests widely include Face Recognition, Action Recognition, Domain Adaptation, Deep Generative Model, and Adversarial Learning. His papers appear at top-tier venues such as Computer Vision and Pattern Recognition, International Conference on Computer Vision, International Conference on Pattern Recognition, and Neurocomputing Journal. 
He is also a reviewer of top-tier journals and conferences including IEEE Transaction on Pattern Analysis and Machine Intelligence, 
IEEE Transaction on Image Processing,
IEEE Transactions on Circuits and Systems for Video Technology,
IEEE Transactions on Artificial Intelligence,
Journal of Computers Environment and Urban Systems, 
IEEE Access, 
Computer Vision and Pattern Recognition,
European Conference on Computer Vision,
International Conference on Computer Vision,
Asian Conference on Computer Vision,
Winter Conference on Applications of Computer Vision, 
International Conference on Pattern Recognition.
\end{IEEEbiography}

\begin{IEEEbiography}[{\includegraphics[width=1in,height=1.25in,clip,keepaspectratio]{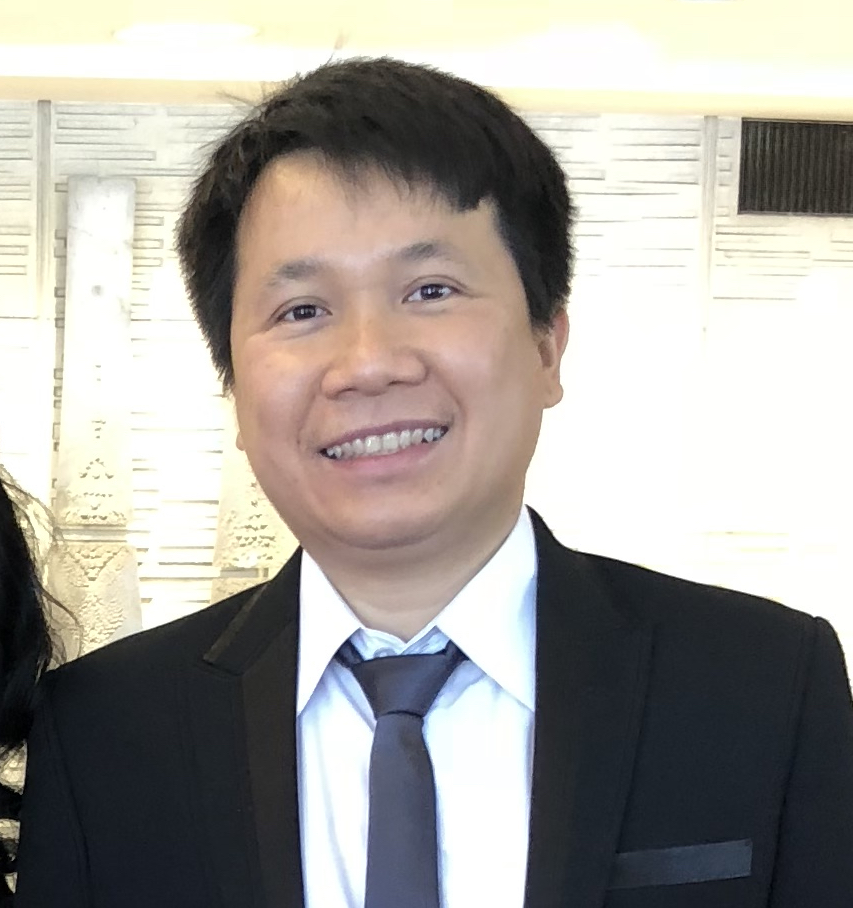}}]{Chi Nhan Duong}
is currently a Senior Technical Staff and having research collaborations with both Computer Vision and Image Understanding (CVIU) Lab, University of Arkansas, USA and Concordia University, Montreal, Canada. He had been a Research Associate in Cylab Biometrics Center at Carnegie Mellon University (CMU), USA since September 2016. He received his Ph.D. degree in Computer Science with the Department of Computer Science and Software Engineering, Concordia University, Montreal, Canada. He was an Intern with National Institute of Informatics, Tokyo Japan in 2012. He received his B.S. and M.Sc. degrees in Computer Science from the Department of Computer Science, Faculty of Information Technology, University of Science, Ho Chi Minh City, Vietnam, in 2008 and 2012, respectively. His research interests include Deep Generative Models, Face Recognition in surveillance environments, Face Aging in images and videos, Biometrics, and Digital Image Processing, and Digital Image Processing (denoising, inpainting and super-resolution). He is currently a reviewer of several top-tier journals including IEEE Transaction on Pattern Analysis and Machine Intelligence (TPAMI), IEEE Transaction on Image Processing (TIP), Journal of Signal Processing, Journal of Pattern Recognition, Journal of Pattern Recognition Letters. He is also recognized as an outstanding reviewer of several top-tier conferences such as The IEEE Computer Vision and Pattern Recognition (CVPR), International Conference on Computer Vision (ICCV), European Conference On Computer Vision (ECCV), Conference on Neural Information Processing Systems (NeurIPS), International Conference on Learning Representations (ICLR) and the AAAI Conference on Artificial Intelligence.
He is also a Program Committee Member of Precognition: Seeing through the Future, CVPR.
\end{IEEEbiography}

\begin{IEEEbiography}[{\includegraphics[width=1in,height=1.25in,clip,keepaspectratio]{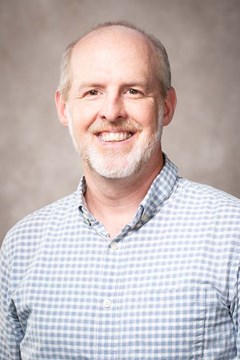}}]{Ashley Dowling} is currently a Professor in the Department of Entomology and Plant Pathology at the University of Arkansas. He is serving as Editor-in-Chief of the International Journal of Acarology. His research interests focus on biodiversity, evolutionary biology, and ecology of insects and other arthropods. He has coauthored 80+ papers in journals on these topics and trained more than 20 graduate students. 
\end{IEEEbiography}

\begin{IEEEbiography}[{\includegraphics[width=1in,height=1.25in,clip,keepaspectratio]{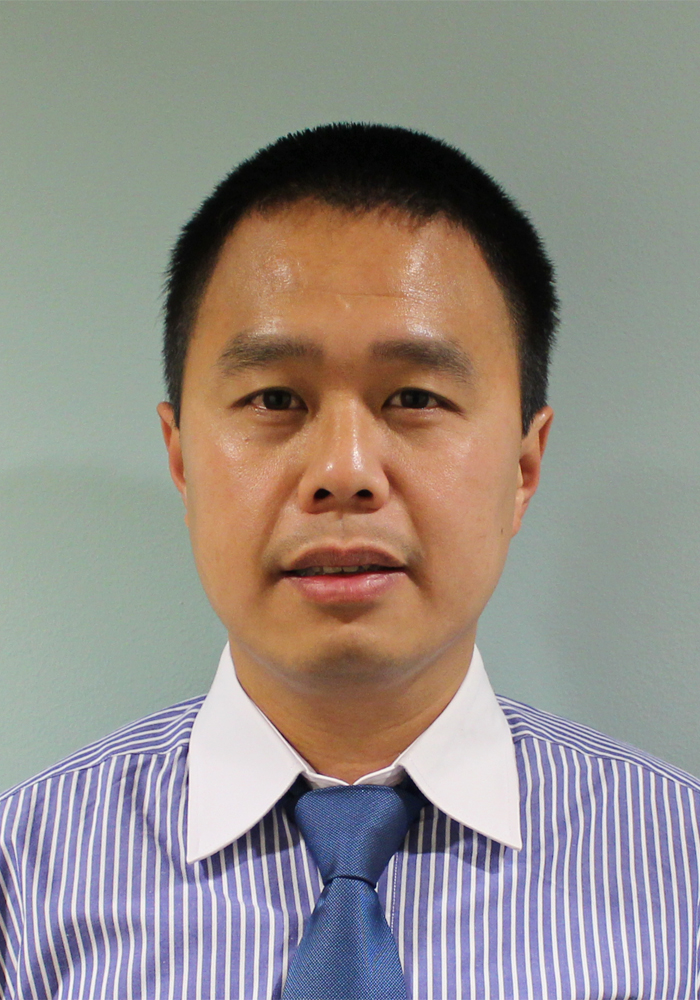}}]{Son Lam Phung}
(Senior Member, IEEE) received the B.Eng. (Hons.) and Ph.D. degrees in computer engineering from Edith Cowan University, Australia, in 1999 and 2003, respectively. He was invited as a Visiting Senior Research Scientist at VinAI and VinFAST, from 2020 to 2021. He is currently a Professor at the University of Wollongong. He has published over 130 papers in journals and international conferences. He has served as the Chief Investigator for over 16 research projects funded by government agencies (research, defense, intelligence, foreign affairs, and trade) and industry. His research interests include image and signal processing, neural networks, pattern recognition, and machine learning. He was awarded the University and Faculty Medals, in 2000. He is currently serving as an Associate Editor for IEEE Access and a Section Editor for Sensors/
\end{IEEEbiography}

\begin{IEEEbiography}[{\includegraphics[width=1in,height=1.25in,clip,keepaspectratio]{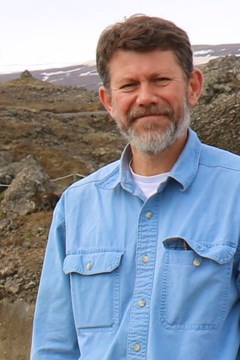}}]{Jackson Cothren} is a professor in the Department of Geosciences at the University of Arkansas. He has experience in photogrammetry, image processing, computer vision, and geodesy. Cothren has a BS in mathematics from the US Air Force Academy, and an MS and a PhD in geodetic science and surveying from the Ohio State University. 
\end{IEEEbiography}

\begin{IEEEbiography}[{\includegraphics[width=1in,height=1.25in,clip,keepaspectratio]{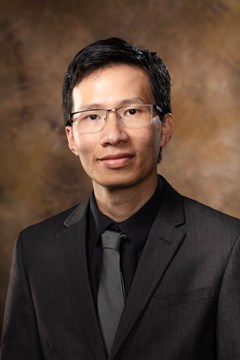}}]{Khoa Luu}
is currently an Assistant Professor and the Director of Computer Vision and Image Understanding (CVIU) Lab
in Department of Computer Science \& Computer Engineering at University of
Arkansas. He is an Area Chair in CVPR 2023. He is also serving as an Associate Editor of IEEE Access journal. He was the Research Project Director in Cylab Biometrics Center at Carnegie Mellon University (CMU), USA. He has received six patents and two best paper awards, and coauthored 150+ papers in conferences and journals. He was a vice chair of Montreal Chapter IEEE SMCS in Canada from September 2009 to March 2011. His research expertise includes Biometrics, Face Recognition, Tracking, Human Behavior Understanding, Scene Understanding, Domain Adaptation, Deep Generative Modeling, Image and Video Processing, Deep Learning, Compressed Sensing and Quantum Machine Learning. 
He is a co-organizer and a chair of CVPR Precognition Workshop in 2019, 2020, 2021, 2022 and 2023; MICCAI Workshop in 2019, 2020 and ICCV Workshop in 2021. He is a PC member of AAAI, ICPRAI in 2020, 2022. He has been an active reviewer for several AI conferences and journals, such as CVPR, ICCV, ECCV, NeurIPS, ICLR, IEEE-TPAMI, IEEE-TIP, IEEE Access, Journal of Pattern Recognition, Journal of Image and Vision Computing, Journal of Signal Processing, and Journal of Intelligence Review.
\end{IEEEbiography}

\end{document}